\pdfoutput=1
\RequirePackage{pdf14}
\documentclass[a4paper,12pt]{book}
\usepackage[utf8]{inputenc}
\usepackage{graphicx}
\usepackage{amsthm}
\usepackage{amsmath}
\usepackage{subcaption}
\usepackage{algorithm}
\usepackage[noend]{algorithmic}
\usepackage{datetime}
\usepackage{epsfig}
\usepackage[colorinlistoftodos,shadow]{todonotes}
\usepackage{xcolor,soul}
\usepackage{ctable} 

\newtheorem*{probdef}{Problem Definition}
\DeclareMathOperator*{\argmax}{arg\,max}

\newcommand{\AUTHOR}{Manuel R. Ciosici}
\newcommand{\TITLE}{Improving Quality of Hierarchical \\Clustering for Large Data 
	Series}

\usepackage[
	style=numeric,
	citestyle=numeric-comp,
	natbib=true,
	backend=bibtex,
	firstinits=true
]{biblatex}
\DeclareNameAlias{default}{last-first}

\AtBeginBibliography{%
}

\DeclareFieldFormat
  [article,inbook,incollection,inproceedings,patent,thesis,unpublished]
  {title}{#1}

\renewbibmacro{in:}{%
  \ifentrytype{article}{%
  }{%
    \printtext{\bibstring{in}\intitlepunct}%
  }%
}

\renewbibmacro*{volume+number+eid}{%
  \printfield{volume}%
  \setunit*{\addcomma\space}%
  \printfield{number}%
  \setunit{\addcomma\space}%
  \printfield{eid}}

\DeclareFieldFormat{pages}{#1}

\renewbibmacro*{publisher+location+date}{%
  \printlist{publisher}%
  \setunit*{\addcomma\space}%
  \printlist{location}%
  \setunit*{\addcomma\space}%
  \usebibmacro{date}%
  \newunit}

\newcommand{\cl}[1]{C_k(#1)}
\newcommand{\p}[1]{p(#1)}
\newcommand{\ALLSAMENOSPACE}{ALLSAME}
\newcommand{\ALLSAME}{\ALLSAMENOSPACE\ }
\newcommand{\RESORTNOSPACE}{RESORT}
\newcommand{\RESORT}{\RESORTNOSPACE\ }
\newcommand{\BROWNNW}{\BROWNNWNOSPACE\ }
\newcommand{\BROWNNWNOSPACE}{BROWN\_NW}
\begin{document}

\pagestyle{empty} 
\pagenumbering{roman} 
\vspace*{\fill}\noindent{\rule{\linewidth}{1mm}\\[4ex]
	{\Huge\sf \TITLE}\\[2ex]
	{\huge\sf \AUTHOR}\\[2ex]
	\noindent\rule{\linewidth}{1mm}\\[4ex]
	\noindent{\Large\sf Master's Thesis, Computer Science\\[1ex] 
		December\ 2015  \\[1ex] Advisors: Ira Assent and Sean 
		Chester\\[15ex]}\\[\fill]}
\epsfig{file=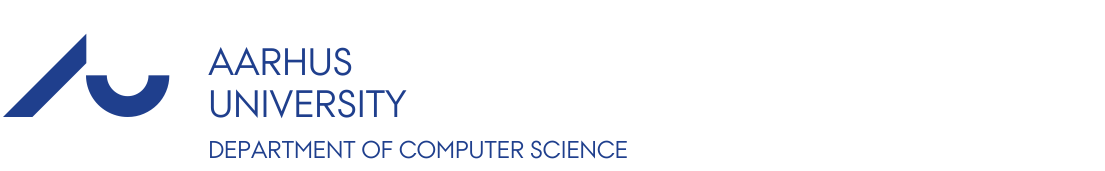}\clearpage


\pagestyle{plain}
\chapter*{Abstract}
\addcontentsline{toc}{chapter}{Abstract}

Brown clustering is a hard, hierarchical, bottom-up clustering of words in a 
vocabulary. Words are assigned to clusters based on their usage pattern in a 
given corpus. The resulting clusters and hierarchical structure can be used in 
constructing \emph{class-based language models} and for generating features to 
be used in natural language processing (NLP) tasks. Because of its high computational 
cost, the most-used version of Brown clustering is a greedy algorithm that uses 
a window to restrict its search space. Like other clustering algorithms, Brown 
clustering finds a sub-optimal, but nonetheless effective, mapping of words to 
clusters. Because of its ability to produce high-quality, human-understandable 
cluster, Brown clustering has seen high uptake the NLP research community where 
it is used in the preprocessing and feature generation steps.

Very little research has been done towards improving the quality of Brown 
clusters, despite the greedy and heuristic nature of the algorithm. 
The approaches tried so far have focused on: studying the effect of 
the initialisation in a similar algorithm (the \emph{Exchange 
Algorithm}); tuning the parameters used to define the desired 
number of clusters and the behaviour of the algorithm; and including a separate parameter to differentiate the 
\emph{window} from the desired number of clusters. However, some of these 
approaches have not yielded significant improvements in cluster quality.

In this thesis, a close analysis of the Brown algorithm is provided, revealing 
important under-specifications and weaknesses in the original algorithm. These 
have serious effects on cluster quality and reproducibility of research using 
Brown clustering. In the second part of the thesis, two 
modifications are proposed. Together, these improve the way in which the 
heuristics are actually applied, improve the quality of Brown clusters and 
provide consistent results. Finally, a thorough evaluation is performed, 
considering both the optimization criterion of Brown clustering and the 
performance of the resulting class-based language models.

\chapter*{Acknowledgments}
\addcontentsline{toc}{chapter}{Acknowledgments}

This thesis was a large project and I could not have done it without help. I 
would like to thank Ira Assent and Sean Chester for providing valuable 
feedback, guidance and access to the computational power necessary for the 
experiments in this thesis. I would also like to thank Leon Derczynski for his 
feedback and for giving me access to extra computational power when it seemed 
like my experiments might not finish on time. Finally, I would like to thank 
Jan Neerbek for letting me bounce ideas off him.

\vspace{2ex}
\begin{flushright}
	\emph{\AUTHOR,}\\
	\emph{Aarhus, December, 2015.}
\end{flushright}
\tableofcontents
\listoffigures
\addcontentsline{toc}{chapter}{List of Figures}
\listoftables
\addcontentsline{toc}{chapter}{List of Tables}
\listofalgorithms
\addcontentsline{toc}{chapter}{List of Algorithms}

\mainmatter
\chapter{Introduction}

A wide variety of computer applications, such as speech recognition, spell 
checking, and predictive text (assisting text input by suggesting next most 
likely words) and machine translation, require computer systems that have 
knowledge of human languages. More specifically, given half a sentence or 
query, they require the ability to predict what word might come next.

Human-spoken 
languages have human-understandable grammars to help people learn the 
language and 
agree on how it should be written and spoken. However, the 
grammars used by humans to learn languages are not suited for computers. 
They make use of what we call \textit{common sense}, which computers severely 
lack. This becomes especially problematic for computers where natural 
language is unclear and common sense is necessary for a proper 
understanding. For example, a formulation such as ``Was good god 
corroded'' might seem to be allowed in English since word order respects 
the Grammar of English. Common sense, though, tells us that because the 
sentence doesn't make sense, it is most likely an error. A computer would 
not be able to tell this as there is no way to express the relationships 
between every possible noun and every possible adjective in English.

Computer scientists have made various efforts to 
develop grammars that computers can understand. \emph{Context Free Grammars} are 
among the most powerful tools trying to address this issue, and were a 
foundation 
stone in developing today's advanced programming languages. Context 
Free Languages are formed of a series of terminal states, representing the 
smallest elements of a language (i.e. words), a list of non-terminal states 
and rules for how to decompose the non-terminal states either into more 
non-terminal states, or into terminal states. By using this system one can 
describe what sort of constructs can appear in contrived languages, such as 
programming languages. However, Context 
Free Grammars do not apply well to natural languages for various reasons, 
such as language inconsistencies, which we humans are quite happy with.  
The limitations of Context Free Grammars have led researchers to 
explore ways to derive models of human languages using automated 
learning techniques. \emph{Statistical Language Models} are a class of 
techniques probabilistically modeling the use of a language. The models 
normally learn by analyzing large corpora of text and inferring what words 
are most likely to follow which other words. For example, given the 
construct \emph{How are}, it is quite easy to imagine that the next most 
likely word is from the set \{\emph{you}, \emph{they}, \emph{the}\}, rather than 
\{\emph{antelopes}, \emph{dwarfs}, \emph{boxes}\}. In normal use of 
English the questions \emph{How are you?} and 
\emph{How are the \dots} are far more common than questions starting 
with \emph{How are antelopes\dots}
 Statistical language models
are 
especially interesting because they generally require no linguistic 
knowledge. A good statistical language modeling 
algorithm can be just as easily applied to French as it can to English; all it 
needs is a sufficiently large corpus of French text from which to learn.

Statistical language models work because they learn to estimate the flow of 
words in a specific language based on its use. The models make some 
assumptions about the structure of 
language 
and learn from the usage observed in a corpus of the language. For 
example, the class of statistical language models dealt with in this thesis 
makes the assumption that the probability of the next word depends only 
on the previous $n$ words, where $n$ is generally a small number like 1 or 
3. This assumption can be easily proven not to hold even with simple 
sentences: In the sentence, \emph{The defendant, said the prosecutor, is 
not guilty}, the word \emph{is} is obviously dependent on the second word 
in the sentence, which is at a distance of 4 words, so any model assuming 
that less than the 4 previous words can influence the next word will not be 
able to model the sentence properly. While such assumptions can easily be 
proven wrong, models based on them work quite well in 
practice. Unfortunately, not all terms in any language 
are used with the same frequency. Low usage of some constructs 
in a corpus creates difficulties for models trying to learn to estimate 
language usage. For example, because this thesis is not on a topic in 
Biology, nouns denoting animals are used sparingly which will cause 
language models trained on the text of this thesis to have problems 
predicting what words should follow \emph{dog} or \emph{elephant}, but 
will give the language model a pretty good idea that after the word 
\emph{language}, we should see the word \emph{model}. The problem of 
reliably predicting words from a corpus is caused by data sparsity. 

The data sparsity problem manifests itself through brittle models. By 
brittle, we mean that a model 
doesn't generalize well. For example, a model 
trained on texts used in news stories and which performs quite well when 
asked to predict text written in the style of a news story, often 
turns out to perform badly when asked to assist in predicting email 
content. \emph{Class-based models} attempt to improve generalization of 
language models at the expense of accuracy. Word classes (or word 
clusters) are groupings of words, such that each group contains words that 
have similar usage in the target language. By grouping words together into 
classes (or clusters), class-based language models can use the occurrences 
of all words in a class to derive a better estimate of how often the class 
occurs. Class-based language models are less accurate, as they are less 
expressive than their non-class-based counterparts. They are also crucially 
dependent on algorithms that create good word classes. It could 
be highly rewarding for all the applications mentioned earlier in this 
chapter to develop algorithms that are better at assigning words to word 
clusters (word classes).

Brown clustering is an assignment of words from a vocabulary $V$ to 
clusters (word equivalence classes) in a hierarchical clustering. A 
hierarchical clustering is a structure similar to a pyramid. Each 
level contains clusters that represent the merging of two clusters on a 
lower level. The lowest level is formed of clusters containing a single word. 
Word assignment to clusters is based on word probability distribution 
within a corpus of text consisting of all words in the vocabulary $V$. Brown 
clustering is a hard clustering, meaning that every word can only be part of 
a single cluster at every level.
Brown clustering can be used as part of a language model, in tasks such as 
spell checking and speech recognition, or as a source of text features in tasks 
such as named entity recognition or machine translation.

The Brown clustering algorithm takes as input a corpus of text composed of 
words $w_1$, $w_2$, $w_3$, \dots, $w_n$ and a parameter $C$ for the number of 
clusters into which the words should be grouped. A bottom-up approach is used to perform
hierarchical clustering. Each word $w_i$ is 
first allocated to its own cluster $C_1(i)$, after which clusters are 
iteratively merged, two at a time, until only one cluster remains, in effect 
producing the hierarchical clustering. The clusters $C_k(i)$ and $C_k(j)$ to 
be merged at step $k+1$ are chosen from among the clusters resulting 
from step $k$ such as to minimize an information theoretical 
measure defined in one of the following chapters. For example, 
the words \emph{Monday} and \emph{Tuesday}, which are generally used in 
the same way, could be merged into a single cluster.

\section{Contributions}
Brown clustering has seen wide uptake in the research community and is used in 
current state-of-the-art algorithms for Named Entity Recognition and Part of 
Speech tagging, as shown later, in Section~\ref{sec:applications_brown}. As 
shown by \citet{brown1992class}, Brown clustering produces nice word clusters. 
However, there is very little research on whether those clusters can be 
dramatically improved. Besides this, there is a lot of ambiguity in the 
definition of Brown clustering which leads to implementations acting more like 
interpretations. In this thesis I challenge the ambiguity in Brown clustering's 
definition and show how it leads to unpredictable results. I also challenge 
some generally held assumptions on the quality of windowed Brown clustering 
against non-windowed Brown clustering. Finally, I propose two modifications to 
the windowed Brown clustering \ALLSAME and \RESORT that aim to make results 
repeatable and improve the quality of resulting clusters

\section{Overview of following chapters}
In Chapter~\ref{chapter_background}, I provide some background on Brown 
clustering, I motivate the use of language models and cover the differences 
between class-based language models and regular language models. I also briefly 
present a few applications of Brown clustering. I cover related research on 
Brown clustering in Chapter~\ref{chapter_related_work}, especially the limited 
work on improving the cluster quality. Chapter~\ref{chapter_analysis} 
contains a thorough analysis of the 
Brown algorithm illustrating weaknesses in the algorithm's original 
specification. I propose two modifications to the windowed Brown clustering 
algorithm in Chapter~\ref{chapter_improvements}, followed by thorough empirical 
evaluation in Chapter~\ref{chapter_experiments}. Finally,
Chapter~\ref{chapter_conclusion} concludes this thesis and presents a few 
directions for future work on Brown clustering.
\chapter{Background}
\label{chapter_background}
This chapter will provide a short introduction to statistical language models: 
what they are and their various kinds (Section~\ref{sec:background_models}). 
The chapter will then introduce 
class-based language models (Section~\ref{sec:background_class_based}), 
what issues they attempt to address. Section~\ref{sec:background_applications} 
will present various applications of statistical language models and of Brown 
clustering.

\section{Statistical language models}
\label{sec:background_models}
A \emph{statistical language model} (\cite{bahl1983maximum, 
brown1992class, ney1994structuring}) is an application of Markov Models 
to natural languages. A \emph{Markov Model} is a stochastic model used to 
model the sequences of transitions in a system of states. When Markov 
Models are used in language models the various words of a vocabulary are 
considered to be the states. Transitions between states are sequences of 
two words that can appear consecutively in the language. Language models 
attempt to estimate the transition probabilities between words in a 
vocabulary. A Markov Model assumes the \emph{Markov Property} which 
requires that future states depend only on the present state. Let $S$ be a 
sequence of text from a language $L$, where $S = w_1, w_2, \dots, w_n$ 
and $w_k$ denotes the $k$'th word in $S$. A {\em bigram language model} for $L$ that 
assumes the Markov Property will assume that the next word ($w_k$) to 
appear in $S$ depends only on the current word ($w_{k-1}$). Formally:
\begin{align}
	\p{w_k | w_1, w_2, \dots w_{k-1}} = \p{w_k | w_{k-1}}
\end{align}
This kind of Markov Model is called a \emph{first order Markov Model}. The model can be extended so that the probability of word $w_k$ depends both on $w_{k-1}$ and $w_{k-2}$, thus creating a \emph{second order Markov Model}:
\begin{align}
\p{w_k | w_1, w_2, \dots w_{k-1}} = \p{w_k | w_{k-1}, w_{k-2}}
\end{align}

The second order Markov Model uses tri-grams (groups of tree words 
appearing consecutively). This is opposed to the first order Markov model 
which uses bi-grams (pairs of words appears after each other in the 
corpus). We can generalize to \emph{higher order Markov models} by using 
\emph{n-grams}:
\begin{align}
\p{w_k} = \displaystyle\prod_{i=1}^{n} \p{w_i | w_1 \dots w_{i-1}}
\end{align}

It is easy to find examples of real-world language use where the Markov 
Property doesn't hold. In the sentence: \emph{They, Angela and 
Martin, are always late}, the word \emph{are} is determined by the first 
word in the sentence (\emph{They}), which 4 words away. It is first with a 
fourth order Markov Model that we would be able to model this 
dependence. However, one could make the iteration of names longer, making it 
practically impossible to use very high order Markov Models. Even though these 
limitations exist in Markov Models, lower order models still manage to do a 
good job at capturing language usage, which is why they are widely used.

Markov models are derived by computing probabilities based on a given 
sample of observations. In the case of language models, the sample of 
observations is a corpus of text. The vocabulary size defines the number of 
parameters to be estimated. For a first order Markov model over a 
vocabulary $V$ of size $|V|$ there are $|V|^2$ parameters (i.e., transitions) to estimate. We 
need $|V|$ probabilities estimated for each of the $|V|$ words in the 
vocabulary.\footnote{One gets $|V|^2$ instead of $|V| (|V|-1)$ because it is not generally true that that word just used cannot be followed by itself.} For a second order Markov model the number of parameters 
becomes $|V|^3$. A language model built even for a small number of 
the words in a language must estimate a large number of parameters. For 
example, a model on a vocabulary of size $20\ 000$ must 
estimate $400\ 000\ 000$ parameters when using a bi-gram model and 
$8\ 000\ 000\ 000\ 000$ (8 trillion) parameters when using a tri-gram 
model.

Since a statistical language model does not have a concept of gender, 
grammatical number or case, the vocabulary of a language model 
corresponds to a much smaller number of language words. For example, a 
language model for English would consider the words \textit{precedented} 
and \textit{unprecedented} as separate, while a language model for French 
would also consider the words \textit{petit}, \textit{petite}, and \textit{petits} as separate 
words since adjectives in French take the gender and grammatical number 
of the noun they describe. Similarly, a statistical language model would not be 
able to distinguish between homographs. The words \emph{fawn} (the creature) 
and \emph{fawn} the verb would be considered the same word, even though they 
define different concepts.

\section{Class-based statistical language models}
\label{sec:background_class_based}
 Besides the large number of parameters to estimate we also need a corpus 
 that is a representative sample of all transitions in order to derive an 
 accurate language model. However, in written language, some expressions 
 and some transitions appear rarely, so large corpora (in the order of 
 millions and billions of words) are needed in order to construct accurate 
 language models even for small vocabularies. \citet{brown1992class} 
 measured a corpus of 365\,893\,263 words of running text with a 
 vocabulary of size 260\,740 to find that there were only 14\,494\,217 
 unique bi-grams (pairs of words appearing one after another). This is only 
 $0.02\%$ of the total of 670 billion possible bi-grams for the given 
 vocabulary size. It is clear that building a reliable bi-gram language model 
 over a vocabulary of this size is not feasible.

\citet{brown1992class} proposed \emph{class-based language models} 
(which I will also refer to as \emph{Brown clustering}) as an alternative to 
word-based statistical language models. In class-based language models, 
words are assigned to \emph{word equivalence classes} (which I will also 
refer to as \emph{word classes} or \emph{word clusters}) based on their 
frequency and pattern of occurrence in a corpus. The underlying idea is 
quite simple: since large vocabularies require estimations of a very large 
number of parameters, Brown clustering reduces the vocabulary size. A 
word equivalence class is a grouping of words that is in effect considered 
a word of its own. Words that tend to be used in similar situations are 
grouped together in the same word cluster (word class). What the language 
model ends up with are groups of words that have a syntactic or semantic 
connection. Figure \ref{fig:example_clusters} shows a subset of five hand-picked 
clusters created by a class-based language model. The last cluster 
consists of country and continent names, which makes sense as these are 
often used similarly.
\begin{figure}
	\begin{verbatim}
	itself himself themselves herself myself 
	until till 
	10 30 12 15 16 17 14 13 11 
	located situated headquartered interred 
	Italy Japan Canada Australia China Europe Ireland England
	\end{verbatim}
	\caption{A hand-picked selection of five word classes derived using Brown clustering. Each row is a cluster.}
	\label{fig:example_clusters}
\end{figure}
 
Brown clustering uses bi-grams (groups of two words following each other 
in the corpus). In class-based language models each word is assigned to a 
single class. The language model then tries to estimate the probabilities of 
transitioning from one class to another. In effect, it becomes a Markov 
Model over the classes of words. Within each class ($c_k$) the language 
model also keeps track of the probability of observing each word ($w_i$) 
that is a member of that class. The probability of the next word $w_k$ 
becomes:
\begin{align}
\label{eq:class-based_language_models}
\p{w_k | w_1, w_2, \dots w_{k-1}} = \p{w_k | c_k}\p{c_k | c_{k-1}}
\end{align}
which states that the probability of the next word is equal to the probability 
of going from the current cluster (word equivalence class) $c_{k-1}$, to the 
class $c_k$ multiplied with the probability of $w_k$ in class $c_k$. However, 
the class-based language model is severely limited in its ability to predict 
the next word in a sequence. Because a language model must predict the most 
likely next word, a class-based language model will be limited to only ever 
predicting the most likely word in each class. Thus, a class-based language 
model resulting from Brown clustering has a prediction vocabulary of only $C$ 
words ($C$ is introduced in the next paragraph). This is part of a conscious 
trade-off between generalization of the model and its accuracy. Later, in 
Chapter \ref{chapter_experiments}, this behavior will also motivate the 
introduction of a novel performance metric for class-based language models.

$C$ is a 
parameter to the class-based model specifying the number of clusters (word 
equivalence classes) to be used; $c_k$ and $c_{k-1}$ are two of the $C$ 
clusters. By using word equivalence classes, the class-based language model 
needs to estimate much fewer parameters. For a first order language model, a 
word-based language model must estimate $|V|^2$ parameters while a class-based 
language model must estimate only $C^2 + V$ parameters: $C^2$ for 
the transitions between clusters and $V$ parameters for the probability of each 
word appearing within the cluster to which it was assigned. For very large values of 
$|V|$ relative to $C$ there are $\approx\frac{|V|}{C}$ fewer parameters to be 
estimated.

Because $C$ is much smaller than $|V|$, class-based language models 
have fewer parameters to estimate, which leads to more reliable parameter 
estimations. In estimating the probability of each class we benefit from the 
occurrences of each word that is a member of the class. However, the lower 
number of model parameters reduces prediction precision. Thus, a 
trade-off is achieved between the complexity of computing a language 
model and the precision of said model. 

In their original work, \citeauthor{brown1992class} propose a greedy
algorithm for computing the class-based language model and deriving word 
clusterings. A greedy algorithm was chosen because of the computational 
complexity of finding an exact solution. In their implementation, the $C$ 
parameter defines an implicit window within which the greedy 
algorithm considers all pairs of clusters ($C^2$ in total) for merging and the 
best candidates are selected. More details on the Brown clustering and other 
algorithms for computing class-based language models are given 
in Chapter \ref{chapter_analysis}. Having defined Brown clustering informally, 
we can now specify its problem definition in a formal manner.
\begin{probdef}[Brown Clustering]
	Given a sequence $W$ of symbols $w_1, w_2, \dots, w_n$ from a 
	vocabulary 
	$V$ and a cluster parameter $C$, with $C \ll |V|$, find the clustering of 
	symbols into $C$ disjoint clusters that maximizes the average mutual 
	information.
\end{probdef}
\begin{figure}[h!]
	\centering
	\includegraphics[width=0.4\textwidth,natwidth=640,natheight=594]{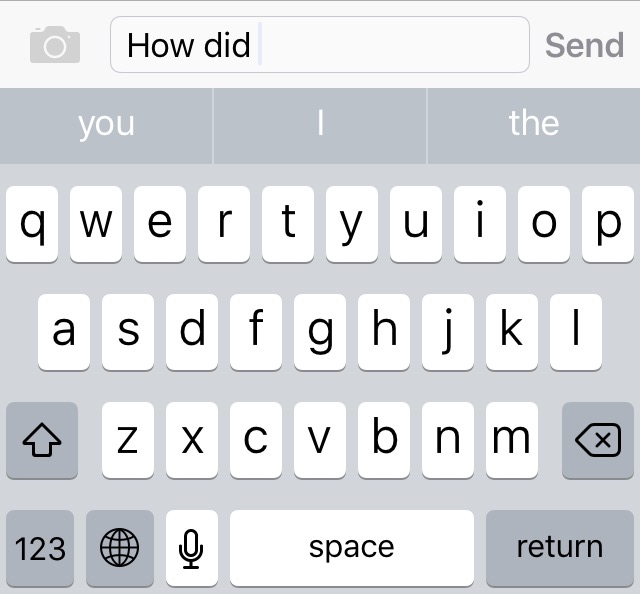}
	\caption{Predictive text input using language models. The underlying 
	language model suggests the words \emph{you}, \emph{I} and 
	\emph{the}.}
	\label{fig:keyboard}
\end{figure}

\section{Applications}
\label{sec:background_applications}

Because Brown clustering results both in a class-based language model and a 
hierarchical clustering, it has applications in two separate areas. The 
following subsections cover these.

\subsection{Applications of Language models}
Statistical language models can be used in applications such as:
\begin{itemize}
 \item Predictive 
text input where a computer system can assist a human typing by 
suggesting the most likely words to be typed next. Figure \ref{fig:keyboard} 
shows a smartphone keyboard assisting the user to type a text message. 
The figure shows a predictive input systems presenting the next most likely 
words. The user can select one of the suggested 
words in order to have it copied into the input field, avoiding spending time 
on typing.
\item A 
speech recognition engine can use language models to improve the 
accuracy of its prediction. For example, in the phrase \textit{We need 
software to recognize speech}, the section \textit{recognize speech} sounds 
almost the same as \textit{wreck a nice beach}. A speech recognition 
engine can use a language model to infer from the context of previous 
words that the speaker is talking about speech recognition rather than 
destroying beaches.
\end{itemize}

\subsection{Applications of Brown clusters}
\label{sec:applications_brown}
Besides assigning words to word equivalence classes, Brown 
clustering can create a hierarchical structure of the clusters, if allowed to 
continue performing merges past the $C$ parameter. The hierarchical 
structure is a binary tree where the $|V|$ leaves are the words of 
vocabulary $V$. All other nodes in the tree represent clusters formed from 
merges of the $|V|$ words. If one labels branches going out of any node 
using 1s and 0s, each cluster obtains a unique 
binary string. The strings can be used in order to provide features for 
\emph{Natural Language Processing} tasks such as \emph{Named Entity 
Recognition}~\cite{miller2004name, Ratinov2009, Turian2010}. Named Entity 
Recognition is the task of identifying and classifying names of persons, 
organizations and places in a given text. Besides Named Entity Recognition, the 
tree resulting from Brown clustering can also be used to generate features for 
\emph{dependency parsing}~\cite{koo2008simple}. \citet{liang2005semi} continued 
on this path and expanded the use of Brown clusters to {\em Chinese word 
segmentation}.

\section{Summary}
This chapter presented statistical language models, an automatic modelling 
method for human language. Statistical language models use a representative 
corpus of text in order to estimate transitions between words in a language. 
However, these models tend to generalize poorly and are difficult to train due 
to the large number of parameters requiring estimation. Class-based language 
models address this by reducing the number of parameters requiring 
estimation. Class-based language models first assign words to classes, and then 
estimate probabilities of transitioning between classes using a corpus text. 
Finally, the chapter covered some of the applications of language models and of 
the hierarchical cluster structure resulting from Brown clustering.

\chapter{Related Work}
\label{chapter_related_work}
This chapter presents a review of research in statistical language modeling 
(Section~\ref{sec:statistical_language_models}) and work on alternative methods 
for building statistical language models (Section~\ref{sec:basic_algorithms}). 
The review then moves on to the rather limited work on trying to 
improve class-based language models (Section~\ref{sec:rel:improvement}) and 
in Section~\ref{sec:rel:similar_work_in_hierarchical_clustering} it provides a 
short discussion of similar work on general agglomerative 
clustering, of which Brown clustering is an instance.

\section{Statistical language models}
\label{sec:statistical_language_models}
Because of the variety of important applications, statistical language 
models have been heavily researched with different 
methods presented in the last few decades. Besides the $n$-gram models, 
of which class-based models are part, there is research 
considering alternative algorithms and methods for building statistical 
language models. Of these I mention models based on random forests, 
recurrent neural networks and methods using decision trees. All of these 
methods are covered by \citet{twodecades} in his survey that presents two 
decades of research on statistical language models. For a comparison of these 
various algorithms, I point the reader to a benchmark article by 
\citet{billionword}.

However, very little research has been done towards improving class-based 
language models. Some work has focused on smoothing language models 
\cite[]{chen1996}, while some researchers such as Peter F. Brown 
have focused on industry application of machine translation and 
speech recognition as attested by a large number of patents filed. For a 
while it was believed that there is no need to improve the 
statistical language modeling techniques, but that focus should be put on 
collecting ever larger training corpora \cite{banko2001scaling}. However, 
more recent research seems to disprove the idea and suggests moving 
focus back to improving the language models 
themselves~\cite{curran2002very}. In the following sections I will review the 
research on Brown 
clustering.

\section{Class-based $n$-gram models}

\subsection{Basic algorithms}
\label{sec:basic_algorithms}
Brown, Desouza and Merger \cite{brown1992class} introduced class-based language 
models in order to 
reduce the number of language model parameters that require estimation, 
in an 
effort to strike a trade-off between generalization power of language models 
and accuracy.
\citeauthor{brown1992class} achieve the reduction by collapsing vocabulary 
words into word equivalence classes and then building a first order Markov 
Model using the word equivalence classes. In order to establish how words are 
mapped to word equivalence classes (word clusters) \citeauthor{brown1992class} 
perform a bottom-up, hard, hierarchical clustering of all words in the 
vocabulary. The original specification of Brown clustering, however, suffers 
from some ambiguity as discussed in Chapter~\ref{chapter_analysis}. Despite 
this and because of the syntactic and semantic meaning of the word 
equivalence classes, Brown clustering became a common tool in Natural Language 
Processing, as described in Section \ref{sec:applications_brown}.
\begin{figure}
	\centering
	\begin{subfigure}[b]{0.65\textwidth}
		\centering
		\includegraphics[width=0.8\textwidth,natwidth=3366,natheight=2716]{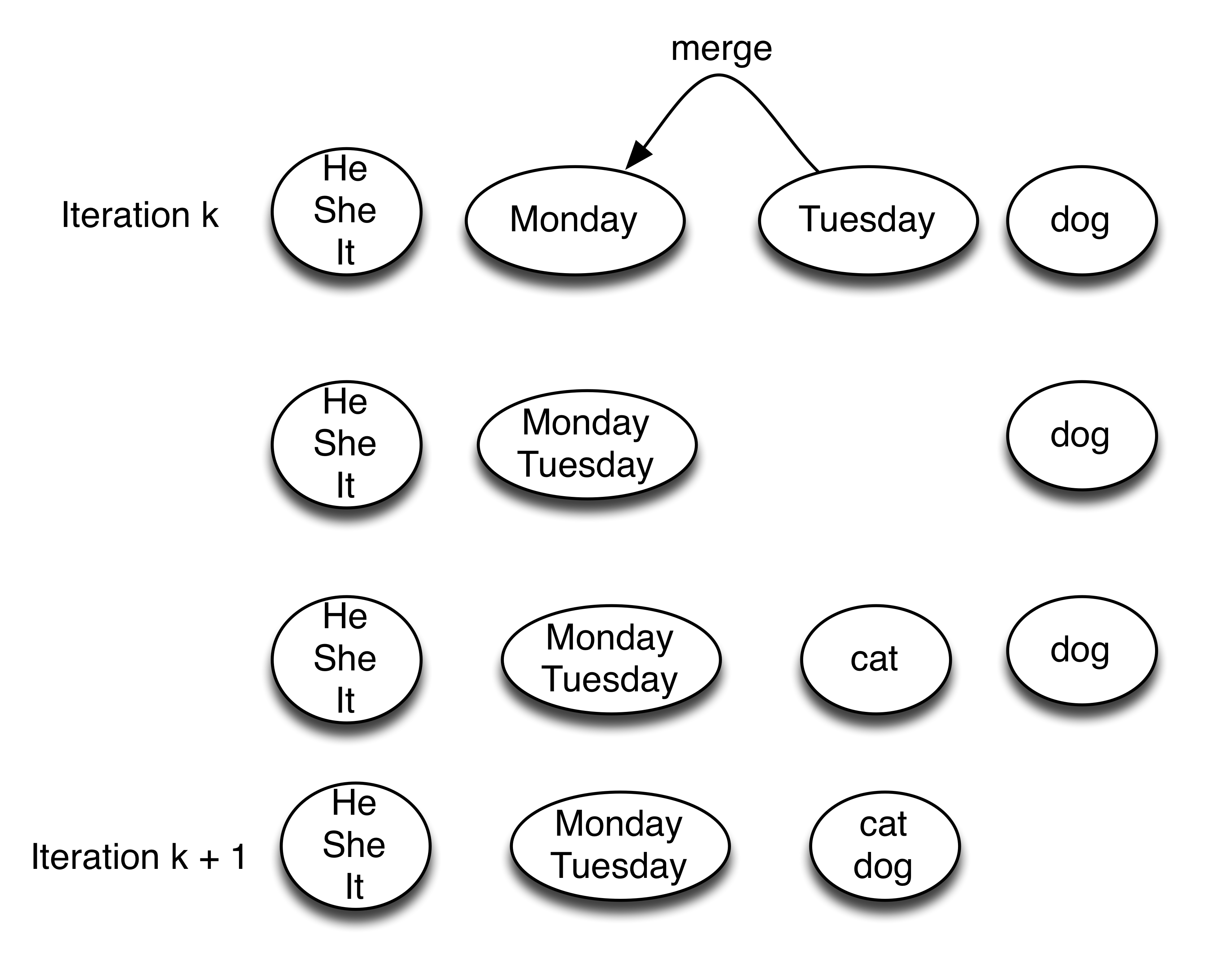}
		\caption{The Brown Algorithm merges two clusters at every iteration 
			step. At every iteration a new word is introduced into the 
			computation 
			window.}
		\label{fig:brown_diagram}
	\end{subfigure}
	\hfill
	\begin{subfigure}[b]{0.3\textwidth}
		\centering
		\begin{tabular}{|c|c|c|}
			\hline  & FROM & TO \\ 
			\hline $w_1$ & $c_1$ & ? \\ 
			\hline $w_2$ & $c_2$ & ? \\ 
			\hline $w_3$ & $c_2$ & ? \\ 
			\hline $w_4$ & $c_2$ & ? \\ 
			\hline $w_5$ & $c_3$ & ? \\ 
			\hline 
		\end{tabular}
		\caption{In each iteration, the Exchange Algorithm must decide where 
		to move each word of every cluster.}
		\label{fig:exchange_diagram}
	\end{subfigure}
	\caption{High-level view of Brown clustering and the Exchange 
	Algorithm}
	\label{fig:algorithms_diagram}
\end{figure}

The exchange algorithm was proposed by \citet{ney1994structuring} as an 
alternative to Brown clustering. The exchange algorithm can be used in 
class-based language models to completely replace \citeauthor{brown1992class}'s 
clustering algorithm. It starts with an initial mapping (covered in section 
\ref{sec:rel:initialization}) from words to classes. It then considers 
for 
each element of every cluster, how the quality of clusters would improve if the 
specific element were to be part of every other cluster. The 
algorithm then performs those moves that lead to the highest increase in 
quality, resulting in a new clustering. Moving a word from one cluster to 
another is called an exchange, hence the algorithm's name. One thing to 
note is that in each iteration of the exchange 
algorithm every word is reassigned to a cluster (which does not necessarily 
have to be a different cluster from its original one). The 
choice of moving 
each word depends only on the clustering at the beginning 
of the iteration and not on any of the move decisions already made during 
the iteration. The algorithm iterates either for a specified number of times 
or until a quality threshold is reached. The stopping criterion is a parameter 
to the algorithm. 

The exchange algorithm is quite different from Brown 
clustering in several ways. Figure \ref{fig:algorithms_diagram} 
shows a high-level comparison between the Brown algorithm and the 
Exchange algorithm. Firstly, the Exchange Algorithm allows for words to 
switch from the cluster they have first been assigned to, unlike Brown 
clustering. At every iteration, the Exchange Algorithm must decide for every 
word in the vocabulary, to which cluster it should be moved to. In contrast, 
the Brown clustering algorithm, at every iteration, performs a non-reversible 
merge of two clusters. The Exchange Algorithm always uses the entire vocabulary 
and set of clusters for every exchange decision. It has no faster, restricted 
version that provides an approximation, like the windowed variant of Brown 
clustering. Secondly, some domain knowledge can be included in the 
exchange algorithm (see section \ref{sec:rel:initialization}). Lastly, the 
Exchange Algorithm does not generate a hierarchy of clusters. Its output 
consists solely of the final clustering.

The exchange 
algorithm might seem similar to the \emph{k-means} \cite{macqueen1967some} 
clustering algorithm, which also iteratively creates clusterings over a 
collection of points. The \emph{k-means} algorithm has the concept of 
\emph{cluster center} and a distance from points to the center of a cluster. 
The exchange algorithm doesn't have \emph{cluster centers}. Its 
optimization goal, the \emph{average mutual information} defined later on, in 
Equation~\ref{eq:ami}, is a function over the interaction of all words, rather 
than a distance function. Brown clustering and the Exchange Algorithm have the 
same optimization goal.

\citet{Martin1998} extended the class-based language models proposed by 
\citeauthor{brown1992class} by defining a version that uses tri-grams as the 
basis for class assignment, in effect constructing a class-based language model 
that uses a second order Markov Model where the probability of a word $w_k$ is
\[ P(w_k | w_1, w_2, \dots w_{k-1}) = P(w_k | c_k)P(c_k | c_{k-1}, c_{k-2})\] 
where $c_{k-1}$ and $c_{k-2}$ are the clusters (word equivalence classes) for 
words $w_{k-1}$ and $w_{k-2}$ respectively. However, \citeauthor{Martin1998} 
do not follow the bottom-up clustering 
method proposed by \citeauthor{brown1992class} Instead, 
\citeauthor{Martin1998} use the exchange algorithm proposed by 
\citet{ney1994structuring}. \citeauthor{Martin1998} classify their method as a 
top-down approach because it starts with the desired number of clusters and 
refines the clusters. However, no hierarchical structure is generated by their 
approach, distinguishing it from Brown clustering and limiting its applications 
to class-based language models only (see Section~\ref{sec:applications_brown} 
for a short discussion on some applications of the hierarchical structure).

\citeauthor{Martin1998} start by assigning the first 
$C-1$ most popular words to individual clusters and the remaining words to 
cluster $C$. They then run the exchange algorithm for a user defined 
number of iterations, or until no more improvements are obtained. The 
optimization goal here is to increase the average mutual information value, 
similar to \citeauthor{brown1992class}'s algorithm. However, 
\citeauthor{Martin1998}'s algorithm maintains a set of $C$ clusters 
over the entire corpus at every iteration. Because of this, 
\citeauthor{Martin1998}'s algorithm can be considered an \emph{anytime 
algorithm}. The Exchange Algorithm stops when it reaches a specified 
number of iterations or 
threshold 
of quality. Because of this it does not construct a hierarchy over the word 
clusters like 
Brown clustering does. The exchange algorithm makes local 
optimizations, and because of this, like 
\citeauthor{brown1992class}'s clustering algorithm, it does not guarantee 
an optimal clustering (i.e. the one that maximizes the amount of mutual 
information).

Another interesting aspect of the Exchange Algorithm is that it can allow 
for the input corpus to be modified. This aspect is not covered in the 
literature. Given clustering $C_1$ of a corpus $T_1$, one can provide the 
Exchange algorithm with clustering $C_1$ as initialization for clustering 
another corpus $T_2$. This technique could be used as a way to provide a 
good initialization for clustering a large corpus. It could also be used to 
improve the clustering of a vocabulary as more corpus data is collected. 
This rather obvious aspect of the Exchange Algorithm is not present in the 
Brown clustering algorithm where, given a higher quality (or longer) version 
of a corpus, one must start clustering from the beginning.

The problem of improving the quality of Brown clustering is a young research 
topic. So far efforts have focused on three main directions: good 
initialization, hyper-parameter tuning and improvements in word 
allocation to classes.

\subsection{Improvement directions}
\label{sec:rel:improvement}
\subsubsection{Effect of initialization}\label{sec:rel:initialization}

\citeauthor{Martin1998} approach the question of improving quality of 
clustering from the view point of selecting a better initialization strategy. The 
algorithms proposed by \citeauthor{Martin1998} and by 
\citeauthor{brown1992class} both initialize by picking the first $C$ most often 
occurring 
words in the corpus. The original papers do not provide a motivation for 
ordering words by frequency. Section \ref{sec:importance_of_ordering_words} 
tries to provide some arguments as to why frequency ordering makes sense and 
what some of its disadvantages are. In the Brown clustering algorithm, ordering 
by frequency leads to the 
clustering process being dominated by high frequency words as they establish 
what the first clusters are. \citeauthor{Martin1998} studied multiple ways 
of initializing the clustering algorithm such as random selection of words for 
initial classes, using the top most frequent words as well as an 
initialization strategy based on Part of Speech classes. In the Part of Speech 
experiment, words were assigned to one of the 33 classes defining parts of 
speech as found in a lexicon. The remaining classes were filled in with the 
most frequent words in the corpus.
 Their experiments 
showed no significant increase in average mutual information, leading them to 
conclude that either word class assignments are independent of initial word 
classes or a better method is still to be found. However, 
\citeauthor{Martin1998} only evaluate the average mutual information on a 
test corpus and do not evaluate on real word natural language 
processing tasks. The test corpus in \citeauthor{Martin1998}'s work is a 
shorter 
text that is completely separated from the corpus used to derive word 
clusterings. This follows the usual approach in Machine Learning, where data is 
separated into a large portion that is used for learning and a smaller one that 
is used exclusively to evaluate performance. The reasoning for such a split is 
to force the learning algorithm to optimize for the test set, without giving it 
the possibility to learn from the test set. In this way, the algorithm's 
generalization is measured, rather than how well it fits the data in the 
learning set.

\subsubsection{Hyper-parameter tuning}
\citet{derczynskitune} investigated the effects of tuning the hyper-parameter 
$C$ (number of desired clusters) on the quality of Brown clustering by 
utilizing the resulting clusters in two classic natural 
language processing tasks: part of speech tagging and named entity 
recognition. \citeauthor{derczynskitune} do not propose a new algorithm, but 
evaluate the effects of various values of $C$ (number of clusters) on clusters 
obtained with \citeauthor{brown1992class}'s original algorithm. They also 
scale the corpus size to see if, for fixed $C$, corpus size increases always 
produce better features.

Their experiments show an increase in cluster quality with an 
increase in corpus size. Similarly, an increase in the number of 
word classes leads to better clustering quality. However, they also show that 
performance tends to plateau and even decrease, when the 
number of clusters increases too much, or the corpus becomes too large, 
leading to noisy word classes. 
\citeauthor{derczynskitune} conclude by recommending researchers try out 
various parameter values for Brown clustering. They also strongly advise 
against the common practice of setting the $C$ parameter to the fixed value of 
1000.

\subsubsection{Improvements in cluster quality}
Another research direction for cluster quality improvement has been explored 
by \citet{chester2015brown}. In their work, \citeauthor{chester2015brown} 
propose the hyper parameter $C$ (number of clusters) be separated from the 
window of data considered for cluster merging. In the original Brown 
algorithm, the $C$ parameter defines an implicit window within which the greedy 
algorithm selects from all pairs of clusters ($C^2$ in total) the 
best candidates for a merge. \citeauthor{chester2015brown} remove this 
restriction by introducing a new parameter $a$ that defines the size of the 
active set to be used for considering cluster mergers. The separation of $C$ 
and $a$ allows for a large search space for cluster mergers (which can 
easily be dealt with by modern computers) without requiring an increase in the 
number of clusters ($C$) that often leads to lower quality clusters as 
mentioned above. \citet{chester2015brown} also developed a different 
method for generation of features from the tree structure resulting from Brown 
clustering by making more use of the average mutual information of the 
clustering. They applied it to Named Entity Recognition, but the method is not 
application specific.

\section{Similar work in hierarchical clustering}
\label{sec:rel:similar_work_in_hierarchical_clustering}
Since the Brown clustering algorithm provides a hierarchical clustering of 
word classes, one should also consider work done on general hierarchical 
algorithms. Information theory has been applied to hierarchical clustering 
\cite{Aghagolzadeh2007, kraskov2009mic, slonim2005information, 
Kraskov2003}. However, these approaches generally rely on assumptions of 
Euclidean space providing some similarity measures that are used at 
various points during the clustering process.

In traditional hierarchical clustering, several approaches have been studied in 
order to improve the quality of clustering. \citet{Salvador2004} proposed a 
method to find the best trade-off between the number of clusters and quality 
according to an 
evaluation metric. The method makes use of a property of graphs, plotting the 
number of clusters versus quality: the sections before and after 
the ideal number of clusters are often approximatively linear. The \textit{L 
method} proposed by \citeauthor{Salvador2004} searches for the point in the 
graph that most exhibits this trait. That point will be selected as the ideal 
number of clusters.

For metric spaces \citet{Dasgupta2005} have proposed a 
hierarchical clustering algorithm with a theoretical cost of at most eight times 
that of the optimal k-clustering. The cost of a clustering is defined as the 
largest radius of its clusters. The algorithm uses the \textit{farthest-first 
	traversal} of a set of points in order to establish an order for points in the 
dataset. The \textit{farthest-first traversal} of a set of points starts by randomly 
selecting a point. All subsequent points are selected so that they are furthest 
away from all the points that have already been selected:
\[
\forall i > 2,\ i = \argmax_i\ \min {d(i, j) : j \in \lbrace 1, 2, 
	\dots i \rbrace }
\]
\citeauthor{Dasgupta2005} construct edges between the points in the 
\textit{farthest-first traversal} by using the concept of granularity. At a 
granularity level of $l$, one can only see points $p$ from the dataset that are at 
least a distance $R_l$ from the first $l$ selected points. The distance $R_l$ 
becomes geometrically smaller as the granularity level increases. The graph 
is iteratively split into connected components. All points in a connected 
component are allocated to the same cluster. The iterative splitting process 
constructs the hierarchical clustering structure.

\citet{balcan2014} proposed a hierarchical clustering algorithm that, given 
that the data to be clustered respects some properties, can work in an 
inductive way. More specifically, given a small subsample of the data to be 
clustered, the algorithm can create a hierarchical clustering structure that 
generalizes to the entire data set. For readers interested in more 
information on hierarchical clustering I would like to point the reader to a 
current survey of such algorithms by \citet{mullner2011modern}.

\section{Summary}
This chapter presented previous research on statistical language models 
(Section~\ref{sec:statistical_language_models}). 
Section~\ref{sec:basic_algorithms} then covered the basic algorithms available 
for computing Brown clusters. A review of the rather limited work on quality 
improvements was presented in Section~\ref{sec:rel:improvement} followed by a 
review of related work on other hierarchical clustering algorithms in 
Section~\ref{sec:rel:similar_work_in_hierarchical_clustering}.

\chapter{Analysis of Brown clustering}
\label{chapter_analysis}

This chapter provides an in depth description of the Brown clustering algorithm 
in Section~\ref{sec:introduction_and_notation}. Following, 
Section~\ref{sec:insights} covers some novel insights into the importance and 
effects of underspecified parts of the Brown clustering algorithm.

\section{Introduction and notation}
\label{sec:introduction_and_notation}
In their work, \citeauthor{brown1992class} favour a less formal 
definition of the Brown clustering algorithm. For example, 
\citeauthor{brown1992class} does not provide a pseudocode description of the 
algorithm. In this section, I will try to present a more formal definition of 
the algorithm following the work of \citet{chester2015brown}. My hope is that 
providing a more formal definition of my changes will help in the effort of 
creating implementations of my proposals and in repeatability of my work.
To my knowledge there are only two openly available implementations of 
the original Brown clustering algorithm: One written in \texttt{C++} by 
\citet{liang2005code}\autocite{liang2005semi} and one in \texttt{python} by 
\citet{heilman2013}. In the following section I will refer to the two 
implementations when discussing details of the Brown clustering algorithm. 
Another thing I should mention is that most of the literature on Brown 
clustering uses the term \emph{token} to denote any element that can be 
clustered. This means both words and punctuation marks (like ``.'', ``-'' or ``!'') are clustered. 
In the rest of this document I will use the term \emph{words} to mean the exact 
same thing as the literature means with the term \emph{token}. I made this 
choice in order to avoid burdening the reader with extra terminology.

Let the input $S$ be a corpus of text with a vocabulary $V$ of size $|V|$. By 
$V[k]$ I will denote the $k$'th word in vocabulary $V$. I will use the term 
\emph{word} rather liberally to denote both actual language words like 
\emph{dog} or \emph{the}, but also punctuation signs like \emph{?} 
(question mark) or \emph{(} (open parenthesis). The definition of words 
used here is case sensitive, so the words \emph{The} and \emph{the} are 
distinct words in the vocabulary $V$. This is meant to capture the cases where 
capitalization indicates completely different concepts. For example \emph{no} 
is the English negation while \emph{No} is shorthand for \emph{number} and {\em Danish} 
is an adjective referring to things originating in Denmark (such as this thesis) whereas {\em danish} is a 
delectable pastry. The 
common case where the capitalized and non-capitalized words denote the same 
concept is trivially handled by Brown clustering as the words will have a 
similar usage pattern (this will become clearer later in the chapter).
 
I will use $C$ for the parameter denoting the number of clusters, $C_k$ to 
denote a clustering which is a collection of $C$ disjoint clusters $C_k(1), 
\dots, C_k(C)$ containing one or more words from the vocabulary $V$ so 
that each word from $V$ is assigned to one cluster. The $k$ subscript 
identifies $C_k$ as being the clustering resulting after $k$ 
merge steps of the Brown clustering algorithm. Formally, 
$$C_k(i) = C_k(j) \implies i = j,\ \ \forall i,\ j$$ 
$$|C_k(i)| \geq 1,\ \ \forall i,\ k $$
$$ \bigcup_i C_k(i) = V, \ \  \forall k$$

$p(\cl{i}, \cl{j})$ will be used to denote the probability that a word in class 
$C_k(i)$ follows a word in class $C_k(j)$. 
If we consider this over all the clusters available, we end up with a matrix of 
co-occurrences, which I will also refer to as \emph{the occurrence matrix}. 
The occurrences will 
be derived from the input 
corpus $S$ by counting. For shorthand, I will use the notation $\p{i, j}$ to 
index into the occurrence matrix at 
row $i$ and column $j$. By $pl(\cl{i})$ and $pr(\cl{i})$ we will denote the 
left and right probabilities of cluster $\cl{i}$. By probability left 
(respectively right) of $\cl{i}$ I mean the probability of any word in 
$C_k(i))$ appearing before (respectively after) any word $m$ in the 
vocabulary $V$ (including itself):
\begin{align}
\label{eq:pl}
pl(i) = \sum_{m = 0}^{|V|} p(i, V[m])
\end{align}
\begin{align}
pr(i) = \sum_{m = 0}^{|V|} p(V[m], i)
\end{align}

Since using $\cl{i}$ to denote the $i$'th cluster can become tedious, I will 
also refer to cluster $\cl{i}$ as simply $i$, where this doesn't create any 
confusion. Therefore, $pl(\cl{i})$ is going to be the same as $pl(i)$.

The \emph{Mutual Information} of two clusters $i$ and $j$ will be denoted by
\begin{align}
\label{eq:mutual_information}
MI(i,j) = p(i,j) \log_2  \frac{p(i,j)}{pl(i)pr(j)}
\end{align}
The \emph{Average Mutual Information} (AMI) of a clustering $C$ is the sum of 
the mutual information of any two clusters $i$ and $j$:
\begin{align}\label{eq:ami}AMI = \sum_{i,j \in C_k}MI(i,j)\end{align} Even 
though it is an average, Equation~\ref{eq:ami} does not contain any divisions 
as each mutual information term is weighed through the use of $p(i,j)$ in 
Equation~\ref{eq:mutual_information}.

A merger of two clusters $\cl{i}$ with $\cl{j}$ with $i < j$ is the operation 
of moving all words from cluster $\cl{j}$ into cluster $\cl{i}$. I will use the 
short notation $i + j$ to denote the merger of cluster $\cl{i}$ with $\cl{j}$. 
Mergers of the form $i 
+ i$ are not allowed. Given this notation, we have the following equations 
for all $i$, $j$, $m$, including $m = i + j$:
\begin{align}
p(i + j,m) = p(i, m) + p(j, m)
\end{align}
\begin{align}\label{eq:plmij}
p(m, i + j) = p(m, i) + p(m, j)
\end{align}
\begin{align}
pl(i + j) = pl(i) + pl(j)
\end{align}
\begin{align}
pr(i + j) = pr(i) + pr(j)
\end{align}
\begin{align}
MI(i + j,m) = p(i + j, m) \log_2  \frac{p(i + j, m)}{pl(i + j)pr(m)}
\end{align}
\begin{align}\label{eq:qmij}
MI(m, i + j) = p(m, i + j) \log_2  \frac{p(m, i + j)}{pl(m)pr(i + j)}
\end{align}

Equations \ref{eq:plmij} and \ref{eq:qmij} are not present in the original 
paper by  \citeauthor{brown1992class}, but they follow the implied idea of 
the algorithm. Being a greedy algorithm, Brown clustering will choose to 
merge those two clusters that lead to the lowest loss in average mutual 
information. \citeauthor{brown1992class} define the loss in average mutual 
information as 
\begin{align}\label{eq:loss}
L(i, j) \equiv AMI - AMI(C_k \setminus \cl{i} \setminus \cl{j} \cup \cl{i+j}), 
\end{align}
where the second term is the average mutual information of the clustering 
$C_k$ where clusters $i$ and $j$ have been merged. Although not 
specified in the original paper, a merge of two clusters doesn't necessarily 
lead to a lower average mutual information. This is the reason for using the 
equivalence symbol ($\equiv$) in Equation \ref{eq:loss}. The simplest 
example is to consider what would happen in the windowed algorithm 
(presented later in this chapter), when we merge two words that have no 
occurrence with any cluster in the window, including each other. Their mutual 
information is 0 (since the nominator in Equation \ref{eq:mutual_information} 
is 0). Similarly, the mutual information terms will be 0 for every pair of  
clusters where one of them is one of the clusters being merged. So, the total 
loss in AMI loss is 0. In practice, the equivalence should be read as an 
equal sign since a minimization of the difference will lead to the clusters 
that have the lowest negative impact on the average mutual information.

The algorithm starts by considering each 
word in the vocabulary as being its own cluster. Given that we have $|V|$ 
words in the vocabulary, and we need $C$ clusters, there are $|V| - C$ 
merges that have to be performed. I only focus here on the merges necessary for 
going from $|V|$ words to $C$ clusters. This will not generate the hierarchical 
structure described in Chapter \ref{chapter_background}. In order to achieve 
that, a total of $|V|$ merges are necessary, $|V|-C$ in order to obtain $C$ 
clusters, and $C$ merges in order to complete the hierarchical structure. The 
class-based language models are already derived after the first $|V|-C$ 
mergers, which is why I only focus on these in this document. The complete 
hierarchy is only necessary for the Natural Language Processing applications 
described in Section \ref{sec:applications_brown}, and, anyway, a straight-forward 
adaptation of the ideas presented here.

Returning to the first $|V|-C$ mergers, at each step we must consider a merge 
of any two pairs of possible clusters. In order to make any merge we need to 
consider $\frac{(|V| - k - 1)^2}{2}$ combinations. We start with $|V|$ 
clusters (one for each word) and after $k$ iterations we have $|V| - k$ 
clusters left, since every iteration merges two clusters. 
There are $(|V| - k - 1)^2$ pairs possible at every iteration since we don't 
allow a cluster $w_k$ to be merged with itself. Performing the 
merge $i + j$ is the same as $j + i$. Because of this, we only need to check 
half of the pairs, hence $\frac{(|V| - k - 1)^2}{2}$ combinations. However, 
checking all these possibilities might take a long time.

\citeauthor{brown1992class} proposed two approaches to address this. 
The first one is a \emph{dynamic programming} approach to the 
clustering problem. Generally, a dynamic programming algorithm splits a 
problem 
into smaller subproblems. When solving each subproblem, the algorithm 
saves the result of the subproblem (along with other relevant data). When 
the algorithm runs with the same subproblem, or a superset of the 
subproblem, the 
cached data structure can be used to either immediately retrieve the correct 
result, or use it to quickly compute a result for the larger problem. Thus, a 
dynamic programming algorithm can provide faster computation at the 
expense of some memory usage. In the Brown clustering algorithm, the 
dynamic programming approach allows for the next pair of clusters to be merged 
and next AMI to be calculated from data at the current iteration, at the 
expense of caching some larger data structures in working memory (Random Access 
Memory), more specifically, the loss in AMI and the quantity of mutual 
information for every hypothetical merge under consideration.

I have chosen not to incorporate the dynamic programming version of Brown 
clustering as the focus of this thesis is not on performance, but on quality 
of results. Non-dynamic programming versions of the algorithm are easier to 
modify.

 The second approach is a windowed algorithm. The windowed algorithm 
 starts by creating one cluster per word, just like the initial Brown clustering 
 algorithm. But, unlike the initial algorithm, the windowed version uses 
 only a subsection of the $|V|$ initial clusters in order to make merge 
 decisions. The window is set to $C + 1$ and at each iteration one merge is 
 performed inside the window, followed by the introduction into the window 
 of a cluster from the ones left outside of the window. Thus, at every 
 iteration there are $C+1$ clusters in the window. However, the algorithm's 
 focusing only on clusters in the window reduces its search space and thus its 
 ability to find optimal solutions. When there are no more 
 clusters to include in the window the algorithm proceeds like the initial 
 Brown algorithm until only one cluster is left. In an attempt to alleviate the 
 effects of the restricted search space, \citet{chester2015brown} proposed that 
 the  window be made into a parameter to the algorithm. This allows windowed 
 Brown to use windows that are larger than $C$, thus making better merge 
 decisions. A pseudocode representation of the windowed Brown clustering 
 algorithm is presented in Algorithm~\ref{alg:windowed_brown}. Lines 1 and 2 
 create the initial $C$ clusters out of the most frequent $C$ words. Lines 3 to 
 11 represent the main algorithm loop. On lines 4~-~5 the merge leading to the 
 lowest loss is performed and the resulting cluster used to replace the cluster 
 with the lower index in the merge. Then, lines 6~-~11 insert 
 another word into the merge window if that is possible, otherwise they remove 
 the extra cluster. Please note that Algorithm~\ref{alg:windowed_brown} does 
 not describe the part of Brown clustering that generates the hierarchical  
 structure as it is not a focus of this thesis.
 
 \begin{algorithm}
 	\caption{The windowed Brown algorithm}
 	\label{alg:windowed_brown}
 	\algsetup{indent=2em}
 	\begin{algorithmic}[1]
 		\REQUIRE{sequence S, number of clusters C}
 		\FOR{$i = 0$ \TO $C+1$}
 		\STATE include word $i$ in window as $C_i$
 		\ENDFOR
 		
 		\FOR{$i = 0$ \TO $|V_S| - C$}
 		\STATE $(C_i, C_j) \leftarrow 
 		best\_merge(C,\ C+1,\ C+1)$
 		
 		\COMMENT{$best\_merge$ is defined in Algorithm 
 			\ref{alg:allsame:best_merge} on page
 			\pageref{alg:allsame:best_merge}}
 		\STATE $C_i \leftarrow C_{i \leftarrow j}$
 		\IF{$i + C + 1 \neq |V_S|$}
 		\STATE include word $V_S[i + C + 1]$ as $C_j$
 		\ELSE
 		\FOR{$k = j$ \TO $C + 1 $}
 		\STATE $C_k = C_{k+1}$
 		\ENDFOR
 		\STATE remove cluster $C_{C}$
 		\ENDIF
 		\ENDFOR
 	\end{algorithmic}
 \end{algorithm}
 
 The two approaches described above, using dynamic programming and using a 
 window, are theoretically complementary. It should be possible to use them 
 together in order to speed up computations and reduce the algorithm run time. 
 However, the mathematical equations covering the dynamic programming version 
 (equation 17 in the paper by \citeauthor{brown1992class}) only cover the 
 non-windowed Brown clustering algorithm. Furthermore, only a textual 
 description is provided for the windowed version. Because of this situation, 
 it is not directly possible to implement a version of Brown clustering taking 
 advantage of both approaches. The biggest challenge in deriving equations to 
 make the dynamic programming version compatible comes from the fact that newly 
 inserted words can have a serious effect on the interaction of clusters in the 
 window. Inserting a new word into the window will always increase the average 
 mutual information as it adds $C$ new terms to the sum in 
 Equation~\ref{eq:ami}. These $C$ newly added terms affect the costs of making 
 any merge.
%
%
%

\section{Novel insights into windowed Brown clustering}
\label{sec:insights}
The windowed algorithm assumes words in the vocabulary are sorted in 
descending order of their frequency. The first $C + 1$ words are taken 
together and the two of them that result in the lowest loss in AMI are 
merged together. Then, the next most often occurring word is included into 
the window until and the process is repeated until all words have been 
assigned to one of the $C$ clusters. However, up to now, researchers have not 
considered how essential parts of the algorithm such as word order and breaking 
ties affect clustering quality. 

\subsection{Importance of ordering words}
\label{sec:importance_of_ordering_words}
In the original algorithm, clusters are seeded by those words with highest 
frequency, and words are included into the window in order of their frequency 
(most frequent first). But, there is no argumentation for sorting words by 
their frequency. Presumably, the hope in selecting the most often occurring words (as 
opposed to the least frequent ones, for example) is that the occurrence matrix 
will be less sparse, more accurately reflecting the cost of merging clusters, 
leading to better merge choices. If the algorithm is to limit its search space 
by using a window, one would want to start by filling the said window with a 
good set of words. 
One would also like to include words into the window in an order that 
would help the restricted algorithm make good merge choices. Some 
work related to good initialization has been covered in Section 
\ref{sec:rel:initialization}. However, that work applies only to the exchange 
algorithm since Brown clustering does not start with an initial clustering of 
all words in the vocabulary.

As mentioned before, 
not all transitions between words (or word equivalence classes) is expected 
to appear in any training corpus. For many clusters $i$ and $j$, the joint 
probability is $\p{i,j} = 0$, which leads to a sparse occurrence matrix. 
Because $\p{i,j}$ is the first 
term in the definition of mutual information (see Equation 
\ref{eq:mutual_information}), for many cluster combinations, the mutual 
information will be zero. For clusters with a mutual information of 
zero, a merger will result either in no loss, or a very small loss. Such a 
merge is very 
tempting for the greedy Brown algorithm. This situation can be exacerbated by the 
order in which words are included in the merge window. 

Let us consider what would 
happen if one were to include in the merge window a word that happens 
not to occur before or after any of the words already inside the merge 
window, but that occurs rather often before words that are 
not inside the merge window. Let us say the word we have inserted is 
\emph{dog} and the words outside the window that \emph{dog} appears 
together with quite often might be adjectives like \emph{bad}, \emph{good} 
or \emph{smelly}. The newly included word (\emph{dog}) will have a mutual 
information of zero with any word already in the window because it does not 
appear together with any of the words in the window. This leads to a zero 
loss in AMI for any merge, so \emph{dog} will be merged with almost any 
cluster in the window, depending on the implementation of Brown 
clustering. We know, however, that \emph{dog} has a positive value of 
mutual information with some words outside the window. When any of 
those words are brought into the window, say \emph{smelly}, their mutual 
information value 
will be higher with the cluster that contains \emph{dog}, because the 
word \emph{dog} is a member of the cluster. In absence of any occurrences 
of words in the window with \emph{dog} it was 
assigned to a pretty much random cluster. We find ourselves in the 
situation where a random cluster has a high mutual 
information with the newly inserted word (\emph{smelly}), thus affecting 
the merge decision at the current step.

\subsection{The window's effect on probabilities}

From the original paper it is not clear how the merge window should affect 
all of the probabilities 
defined in Equations \ref{eq:pl} to \ref{eq:qmij} above. Should $pl$ and 
$pr$ only count occurrences within the window, or the entire corpus? If we 
use the global counting for $pl$ and $pr$, we will get values more 
representative of the actual use of words in the corpus. If we limit them to 
the window, we obtain values that are more representative of the interaction of 
words within the window. When using the values over the entire corpus $pl$ and 
$pr$ have almost the same value (there will be two words for which the 
values will differ by one, namely the first and last word in the corpus). Using 
the locally computed values is computationally more expensive. 
Using the locally computed values, we will also see much larger 
discrepancies 
between $pl$ and $pr$ and this can change the values of mutual 
information since both $pl$ and $pr$ are part of the mutual information 
formula (see Equation \ref{eq:mutual_information}). In both
my implementation and the one by \citeauthor{liang2005code}, the 
original values for $pl$ and $pr$ are used, i.e. they are not limited to only 
occurrences with words in the window.

If we consider a run of windowed Brown over the corpus in Figure 
\ref{fig:example_text:a} with parameter $C = 2$, the initial window will 
contain words \emph{.} (period), \emph{the} and \emph{cats}. With a global 
count for $pl$ and $pr$, the word \emph{the} will have values 5 and 4, 
respectively. With local counts, however, they become $pl=2$ and $pr=0$ as we 
only count bi-grams with words already in the research window. For $pl$, there 
are two instances of the bi-gram \emph{the cats} and for $pr$, there are no 
bi-grams consisting of a word in the window, followed by the word \emph{the}. 
Since $pr$ is 0, any pair consisting of the word \emph{the} on the right side 
will have a mutual information of 0, while all pairs with \emph{the} on the 
left side will have an increased value of mutual information as local $pl$ is 
less than half the value of global $pl$. The locally counted, lower valued, 
$pl$ and $pr$ will increase the cost of any merge involving the word \emph{the}.

\subsection{Handling words with the same rank}
\label{sec:word_order_example}
The algorithm does not specify how words with the same frequency 
should be arranged in the total ordering that dictates their inclusion into 
the window. This can have considerable implications over the merges that 
are performed, depending on how often the equal frequency words appear 
together with words already in the window. 

\begin{table}[]
	\centering
	\caption{Word frequencies}
	\label{tab:word_frequencies}
	\begin{tabular}{lll}
		rank & word    & no. occurrences \\
		1    & .       & 5               \\
		1    & the     & 5               \\
		2    & cats    & 3               \\
		3    & dog     & 2               \\
		3    & likes   & 2               \\
		3    & Alice   & 2               \\
		4    & chased  & 1               \\
		4    & scared  & 1               \\
		4    & ran     & 1               \\
		4    & away    & 1               \\
		4    & sports  & 1				 \\               
	\end{tabular}
\end{table}
\begin{figure}
	\centering
	\begin{subfigure}[b]{0.3\textwidth}
		\begin{verbatim}
		the dog chased the cats .
		the dog scared the cats .
		the cats ran away .
		Alice likes cats .
		Alice likes sports .
		\end{verbatim}
		\caption{}
		\label{fig:example_text:a}
	\end{subfigure}
	\hfill
	\begin{subfigure}[b]{0.3\textwidth}
		\begin{verbatim}
		the dog chased the cats .
		Alice likes cats .
		Alice likes sports .
		the dog scared the cats .
		the cats ran away .
		\end{verbatim}
		\caption{}
		\label{fig:example_text:b}
	\end{subfigure}
	\caption{Two similar corpora with the same vocabulary, frequency count and 
	co-occurrences}
	\label{fig:example_text}
\end{figure}
\begin{figure}
	\centering
	\begin{subfigure}[b]{0.7\textwidth}
		\caption{Clustering with AMI = 1.1411}
		 \begin{verbatim}
		 the likes 
		 . Alice chased ran scared away 
		 cats dog sports 
		 \end{verbatim}
		
		\label{fig:example_clustering:a}
	\end{subfigure}
	\vfill
	\begin{subfigure}[b]{0.7\textwidth}
		\caption{Clustering with AMI = 1.1373}
		 \begin{verbatim}
		 the likes ran 
		 . Alice chased scared 
		 cats dog away sports 
		 \end{verbatim}
		
		\label{fig:example_clustering:b}
	\end{subfigure}
	\vfill
	\begin{subfigure}[b]{0.7\textwidth}
		\caption{Clustering with AMI = 1.1218}
		 \begin{verbatim}
		 the Alice away 
		 . chased ran scared 
		 cats dog likes sports  
		 \end{verbatim}
		
		\label{fig:example_clustering:c}
	\end{subfigure}
	\caption{Various clusterings that can result from using different 
	interpretations of Brown clustering over the two corpora in Figure 
	\ref{fig:example_text}. Each row represents one cluster.}
	\label{fig:example_clustering}
\end{figure}
Consider the corpus in Figure \ref{fig:example_text:a} with word 
frequencies specified in Table \ref{tab:word_frequencies}. If we run the Brown 
clustering algorithm with parameter $C = 3$, it starts by making a window of 
size $4$ and including the $4$ word with highest frequency. The first three 
of these are: 
\emph{.} (period), \emph{the} and \emph{cats}. The fourth most often 
occurring word is difficult to establish. Is it \emph{dog}, \emph{likes} or 
\emph{Alice}? They all have the same number of occurrences. Actually, 
\citet{brown1992class} do not define what should happen in these 
situations. By leaving the behavior unspecified, we can arrive at different 
implementations providing different clusterings. By continuing the example 
in Figure \ref{fig:example_text:a}, I will also show that the order in which 
the aforementioned words are included can have a significant effect on the 
value of average mutual information.

Let us consider what would happen if the corpus is changed from the one in 
Figure \ref{fig:example_text:a} to the one in Figure \ref{fig:example_text:b}. 
Neither the values of $pl$, nor $pr$, will change since we have the same 
words in the vocabulary and they appear the same number of times. The 
occurrence matrix will also stay the same as the word \emph{.} (period) will 
still appear two times before the word \emph{the} and two times before the 
word \emph{Alice}. The two corpora should have similar clusterings. Because of 
the unspecified behavior mentioned above, the order words are included in the 
merge window changes. This leads to different mergers and, subsequently, to 
different final clusterings and values of average mutual information.

In both my implementation, and that of \citeauthor{liang2005code}, the 
clustering and final average mutual information for parameter $C = 3$ 
changes if the corpus is changed between the ones in Figure 
\ref{fig:example_text}. We will notice a change in average mutual 
information between 1.1411 and 1.1218 as my implementation 
switches 
from the clustering in Figure \ref{fig:example_clustering:a} to the one in 
Figure \ref{fig:example_clustering:b} and \citeauthor{liang2005code}'s changes 
from the clustering shown in Figure \ref{fig:example_clustering:c} to the one 
in Figure \ref{fig:example_clustering:b}. At a first glance the change in 
average mutual information seems small, but we should remember that merges 
earlier in the clustering process can have a significant impact on merges 
later in the process. We can see that clustering in Figure 
\ref{fig:example_clustering:a} does not just have the highest average mutual 
information, but is also an assignment that follows more closely with the kind of 
clustering we, as humans, would expect given the meaning and syntactic role of 
words.

There is one final thing to note. The issue presented in this section applies 
not only to sorting words by frequency. It also applies to the case where words 
are sorted by connectivity (the amount of distinct words appearing after each 
word in the vocabulary), or any other sorting for that matter. Actually, this 
issue is not at all specific to Brown clustering and can affect any sort of 
windowed hierarchical clustering algorithm.

\subsection{Handling non-unique minimal mergers}
Another aspect of tie breaking not specified by \citeauthor{brown1992class} is 
how the clustering algorithm should handle cases where there are several pairs 
of clusters that, if merged lead to the same, equal minimum loss in average mutual 
information.

For this, another problem must be addressed first. When do we consider losses 
of average mutual information to be equal? Here, we have to take into account 
the fact that, very often, the difference between the lowest loss and the 
second lowest loss is smaller than 1. And, because the calculation of average 
mutual information involved lots of operations on floating point numbers, as 
clustering progresses, approximation errors become more pronounced.

One approach in breaking this tie could be to perform all merges that result in 
the lowest loss in average mutual information. However, this becomes 
problematic when the pairs of clusters to merge are not disjoint. Another 
approach could be to perform the second lowest loss in AMI and hope that the 
resulting clustering would break the tie.

\section{Summary}
In this chapter I presented an in-depth look at the Brown clustering algorithm. 
I described the three versions of the algorithm originally proposed by 
\citeauthor{brown1992class}: the non-windowed algorithm, the non-windowed 
algorithm using dynamic programming and a windowed version of Brown clustering. 
The second part of the chapter revealed a number of serious issues in the 
definition of all versions of the Brown clustering algorithm. They doesn't take 
into account the effect of word ordering, words with same frequency or ties 
between different lowest loss mergers. Through the use of examples I made an 
argument for the considerable and undesired effect these omissions can have on 
the quality of resulting word clusters.
\chapter{Quality Improvements}
\label{chapter_improvements}

In this chapter, I describe two algorithmic changes to Brown clustering. The 
first one, \RESORTNOSPACE\ (Section~\ref{sec:resort}), is meant to provide 
a more information 
theoretic method of sorting words in a corpus and to be a starting point for 
general discussions about the idea of using a dynamic word order. The 
second modification, \ALLSAMENOSPACE\ (Section~\ref{sec:allsame}), 
aims at addressing some 
undefined algorithm behavior identified in Chapter \ref{chapter_analysis} by 
guiding merge decisions.

\section{Word ordering}
\label{sec:resort}

Section \ref{sec:importance_of_ordering_words} addressed the choice of 
\citeauthor{brown1992class} to sort words by frequency, before inserting 
them into the merge window. In this section, I propose a variation of the 
Brown clustering algorithm that does not maintain a static word order, 
but in which the order words are inserted into the merge window depends on the words 
that are already in the window.

Sorting words can be motivated by the desire to maximize the amount 
of information about words already in the window. Sorting words by frequency is 
one method that aims at achieving this goal. A simple alternative is to sort 
words by their connectivity. That is, for every word $w$, count the number of 
unique words in the vocabulary $V$ that appear before or after $w$ in a given 
corpus. Sorting by connectivity will tend to create a dense occurrence matrix 
in the merge window (low number of zeros), while sorting by frequency will tend 
to have a higher number of zeros, but larger values for the non-zero values. 

I 
propose words be sorted by 
the amount of mutual information they have with words already in the merge 
window. This can be measured by computing the sum of mutual information between 
every word not in the merge window and every cluster in the merge window (see 
Equation \ref{eq:mutual_information} for the formula of mutual information). 
The resorting can be made after every merge or, in order to reduce the 
computational cost, after every $R$ merges.

To give an intuition of the algorithm, let us consider 
the corpus in Figure \ref{fig:text_margrethe}, with word frequencies 
presented in Table \ref{tab:word_frequencies_margrethe}, and a run of the 
resorting algorithm where the desired number of clusters $C$, is set to 3. The 
initial merge window will have a size of $C+1 = 4$ and contain the words: 
\emph{Margrethe}, \emph{to}, \emph{the} and \emph{II}. After the first merge, 
the word \emph{.} (period) will be inserted into the merge window, and the 
remaining words will be resorted. I will now focus only on the 5 words of rank 
3. By computing the amount of mutual information between these words and 
clusters in the merge window, \RESORT can derive a new ranking. We do not need 
to compute the values of mutual information to understand the example. We only 
need to count the number of co-occurrences with words already in the 
merge window. By looking at Figure \ref{fig:text_margrethe}, we can see that 
words \emph{in}, \emph{her} and \emph{heir} have no co-occurrences with any of 
the 5 words already in the window and thus have a mutual information of 0. The 
word \emph{throne} has one co-occurrence (\emph{the throne}) and the word
\emph{Majesty} has two (both of the form \emph{Majesty Margrethe}). Because of 
the higher number of co-occurrences, the word \emph{Majesty} has more mutual 
information with clusters in the merge window and thus, becomes the next word 
to be included into the merge window. In this case, re-sorting words outside 
the window has managed to break the tie between words with the same frequency. 
By first inserting the word \emph{Majesty}, the algorithm has also chosen to 
involve for merge considerations the word it knows most about, relative to the 
clusters already in the window. I ignored all other words in this 
analysis because they either have a mutual information value of 0 (because of 
no co-occurrences), or a small one (because of having only one co-occurrence).
\begin{figure}
		\begin{verbatim}
		Margrethe II was born on the April 16 1940 . 
		Margrethe II succeeded to the throne in 1972 , 
		becoming Her Majesty Margrethe II.  
		Her Majesty Margrethe II only became heir presumptive in 1953
		when changes to the constitution allowed Margrethe to be a 
		legal heir to the throne.
		\end{verbatim}
		\caption{A corpus that can benefit from re-sorting words not already in 
		the merge window.}
		\label{fig:text_margrethe}
\end{figure}
\begin{table}[]
	\centering
	\caption{Word frequencies for the corpus in Figure 			
	\ref{fig:text_margrethe}}
	\label{tab:word_frequencies_margrethe}
	\begin{tabular}{lll}
		rank & word    & no. occurrences \\
		1    & Margrethe       & 4               \\
		1    & to     				& 4               \\
		1    & the    			   & 4               \\
		1    & II				     & 4               \\
		2    & .  					 & 3               \\
		3    & throne   		 & 2               \\
		3    & Majesty  		& 2               \\
		3    & in  					& 2               \\
		3    & her     			   & 2               \\
		3    & heir    			   & 2               \\
		4    & \emph{all other words} & 1               \\         
	\end{tabular}
\end{table}

\begin{algorithm}
	\caption{The \RESORT algorithm}
	\label{alg:resort}
	\algsetup{indent=2em}
	\begin{algorithmic}[1]
		\REQUIRE{sequence S, number of clusters C, interval for resorting R}
		\FOR{$i = 0$ \TO $C+1$}
			\STATE include word $i$ in window as $C_i$
		\ENDFOR
		
		\FOR{$i = 0$ \TO $|V_S| - C$}
			\STATE $(C_i, C_j) \leftarrow 
				best\_merge(C,\ C+1,\ C+1)$
		
		\COMMENT{$best\_merge$ is defined in Algorithm 
			\ref{alg:allsame:best_merge} on page
			\pageref{alg:allsame:best_merge}}
		\STATE $C_i \leftarrow C_{i \leftarrow j}$
		\IF{$i + C + 1 \neq |V_S|$}
			\STATE include word $V_S[i + C + 1]$ as $C_j$
		\ELSE
			\FOR{$k = j$ \TO $C + 1 $}
				\STATE $C_k = C_{k+1}$
			\ENDFOR
			\STATE remove cluster $C_{C}$
		\ENDIF
		\IF{$i > 0$ \AND $i\ \bmod\ R = 0$ \AND $i \neq |V_S| - C$}
			\STATE $resort\_words(C, V_S, V_S[i + C + 2]$
		\ENDIF
		\ENDFOR
	\end{algorithmic}
\end{algorithm}

In Algorithm \ref{alg:resort}, listing the \RESORT approach, the desired number 
of clusters is denoted by 
$C$, while $C_{index}$ is used to denote a specific cluster in the current 
clustering. Lines 1 and 2 insert the first $C+1$ words into the 
merge window. The algorithm then performs the best merge of two clusters 
(lines 4--5) and inserts the next cluster (lines 6--7). On line 5, cluster 
$C_i$ is replaced by $C_{i \leftarrow j}$, the cluster resulting from the 
merger of clusters $C_i$ 
and $C_j$. If there are no more clusters 
left to insert, the algorithm is at the end of its run, so it removes the 
last $C+1$'th cluster in order to obtain the $C$ clusters required (lines 9--11). Lines 12--13 resort all words starting with the next word to be 
inserted into the window. A general approach to resorting is presented in 
the pseudocode. Resorting can be done after each merge, or after every $R$ 
merges. The $R$ parameter allows the user to trade off a better word 
sorting for a shorter run time, if the resorting process is computationally 
slow. The cost of resorting the next words to include depends on the 
chosen sorting function.

The $resort\_words$ function takes as input the current clustering, an 
ordering of words $V_S$ and a pointer $i$ to one of the words in this 
ordering. It 
then resorts all words between $i$ and $|V_S|$. The resorting criteria can 
be any function that can provide a total ordering. I propose that words 
be ordered by the amount of mutual information they have with words 
already in the merge window. I chose the amount of mutual 
information with words in the merge window as a resorting criteria because 
it involves both connectivity and number of occurrences together with 
words already in the merge window. Connectivity is measures the number of 
unique words that appear in a bi-gram with any given word. Frequency measures 
how often a given word appear in a corpus, as shown in the third column of 
Table~\ref{tab:word_frequencies_margrethe}.

\section{Handling words with same ranking value}
\label{proposing_ALLSAME}\label{sec:allsame}
As discussed in Chapter \ref{chapter_analysis}, the Brown clustering algorithm 
does not specify a strategy for dealing with words that have the same number of 
occurrences in the corpus. As I have shown in Section 
\ref{sec:word_order_example} of Chapter \ref{chapter_analysis}, word order 
is an important algorithmic consideration that can affect the final 
clustering. This raises some deeper 
questions about repeatability of scientific work of researchers using Brown 
clustering as input in their work, or as a pre-processing step. For some 
examples of work using Brown clustering as input or for pre-processing, see 
\cite{chester2015brown, koo2008simple, liang2005semi, 
miller2004name, Turian2010}. 
With the current state of things (word order determined ambiguously), 
different implementations of Brown clustering will 
provide different final clusterings with unpredictable fluctuations in quality. 
In the most widely used implementation of Brown clustering, the one by 
\citet{liang2005code}, the order of words with the same number of occurrences 
is defined by the implementation of sorting in the Standard Library of C++. 
However, if using different compilers or different implementations of the 
Standard Library of C++, even \citeauthor{liang2005code}'s implementation could 
provide different clusterings and values of average mutual information, just by 
changing the compilation environment.

The inclusion order of 
words with same number of occurrences must be taken into account and a behavior 
specified. A 
first idea could be to include as many of the words sharing the same number of 
occurrences as possible into the merge 
window. We could use the ideas in the Generalized Brown clustering proposed by 
\citet{chester2015brown} in order to separate the window from the $C$ 
parameter. To remind the reader, \citeauthor{chester2015brown} proposed that 
the window size be separated from the $C$ parameter so that users must specify 
how large the window size should be (though a minimum of $C+1$). The idea is 
that a larger window size should allow the clustering algorithm to make better 
local decisions given that it has more options to choose from.  We could then 
run the original algorithm using $C+1$ clusters and, 
at every inclusion of a word into the merge window, we would 
check if the next word has the same number of occurrences in the corpus. If 
yes, we would also include the next word and continue including either until we 
encounter a word that has a different number of occurrences, or we reach the 
$w$, the window size. That is, we try to take all words with the same number of 
occurrences in at once.

\begin{figure}[h!]
	\centering
	\includegraphics[width=1.0\textwidth,natwidth=1799,natheight=964]{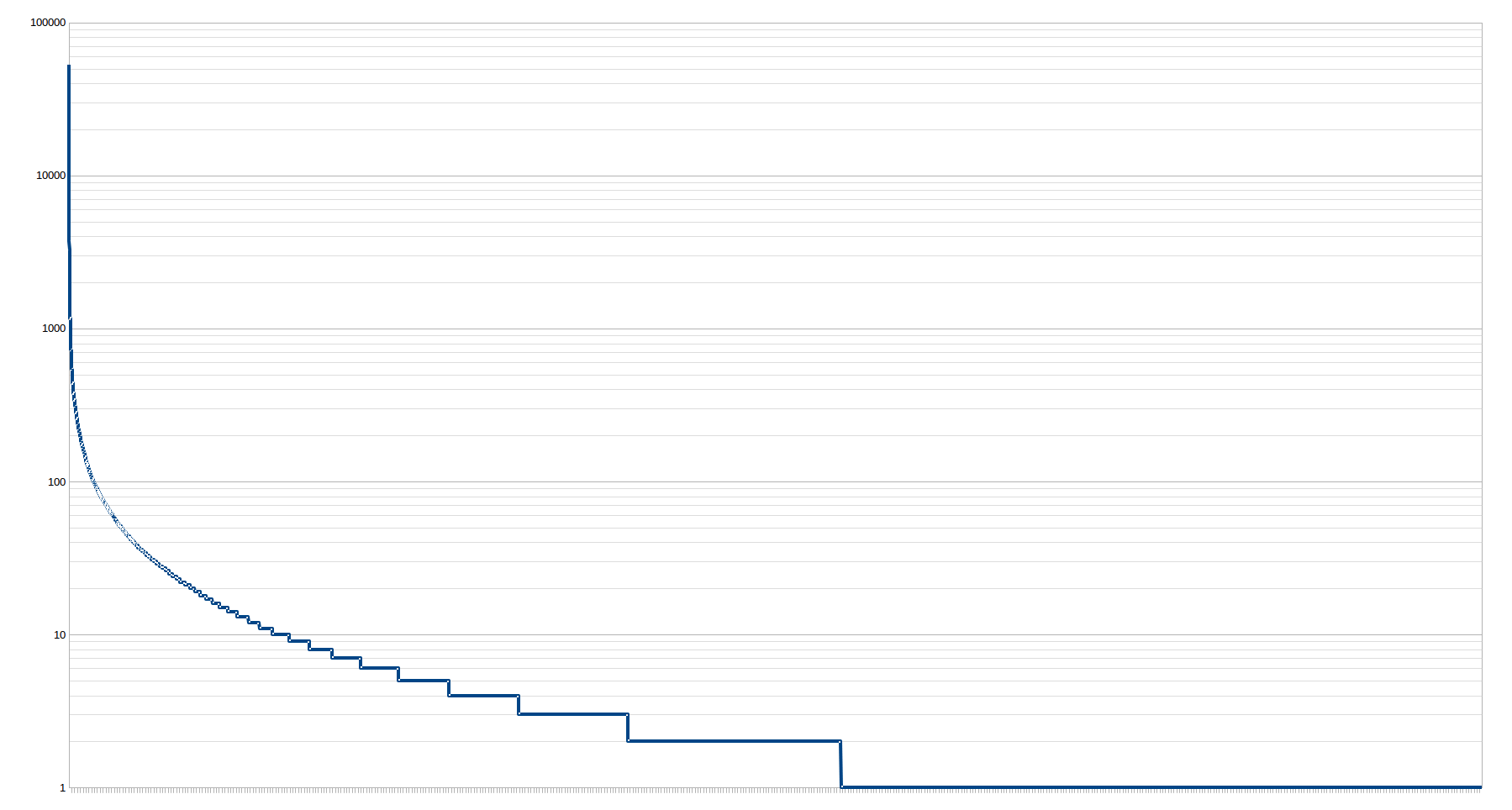}
	\caption{Frequencies of the most frequent 100 words in a corpus of length 
	$1\,000\,000$ and vocabulary of size $56\,000$. A wider step means more 
	words with the same frequency.}
	\label{fig:word_distribution}
\end{figure}

However, there are a few issues with this approach. Because we might reach the 
window size limit before we run out of words with the same number of 
occurrences, we must decide which words to include into the window. A 
simple solution is to increase the window size $w$ so that the limit can never 
be reached. In other words, if we set $w = |V|$, the window will always be 
able to contain all words with the same number of occurrences. Including many 
words at 
a time reveals another problem. As the clustering progresses, the algorithm 
encounters more words with the same number of occurrences (see the flattening 
line in Figure 
\ref{fig:word_distribution}). 
When the algorithm reaches words with a small number of 
occurrences, it encounters more words sharing the same number of 
occurrences and might have to include hundreds, or maybe thousands, of words 
into the merge window. At this point in time, the greedy merge strategy will 
behave quite unexpectedly. Because of
the algorithm's greedy nature, it will always choose to merge those clusters 
that 
lead to the smallest loss in average mutual information. When thousands 
of words with a small number of occurrences are included in the merge window, 
the algorithm will have a tendency to pick one-word-clusters and merge them 
together, since they will generally have a merge cost very close to zero.
 In time, as the window shrinks to approach the size $C$, the one-word-clusters 
 will be merged into two-word-clusters and so on, as they become larger 
 clusters. As the window size $w$ approaches $C$, the loss in average mutual 
 information will increase exponentially since the algorithm will have to merge 
 together larger and larger clusters.

The behavior described above is not present in the original windowed Brown 
clustering because the merge window only receives one word at a time. Almost 
exclusively, the last cluster inserted into the window (which is a 
single-word-cluster) is merged into an already existing cluster. This generally 
happens at a low cost to the average mutual information.

We have seen that including only one word at a time poses the problem of 
choosing an order over words with the same number of occurrences. We have also 
seen that the straight-forward solution of including all words with the same 
number of occurrences can lead to a smaller average mutual information. I 
propose we do not include any word at all, in the method I call 
\ALLSAMENOSPACE. 

\begin{figure}[h!]
	\centering
	\includegraphics[width=0.6\textwidth,natwidth=2516,natheight=1516]{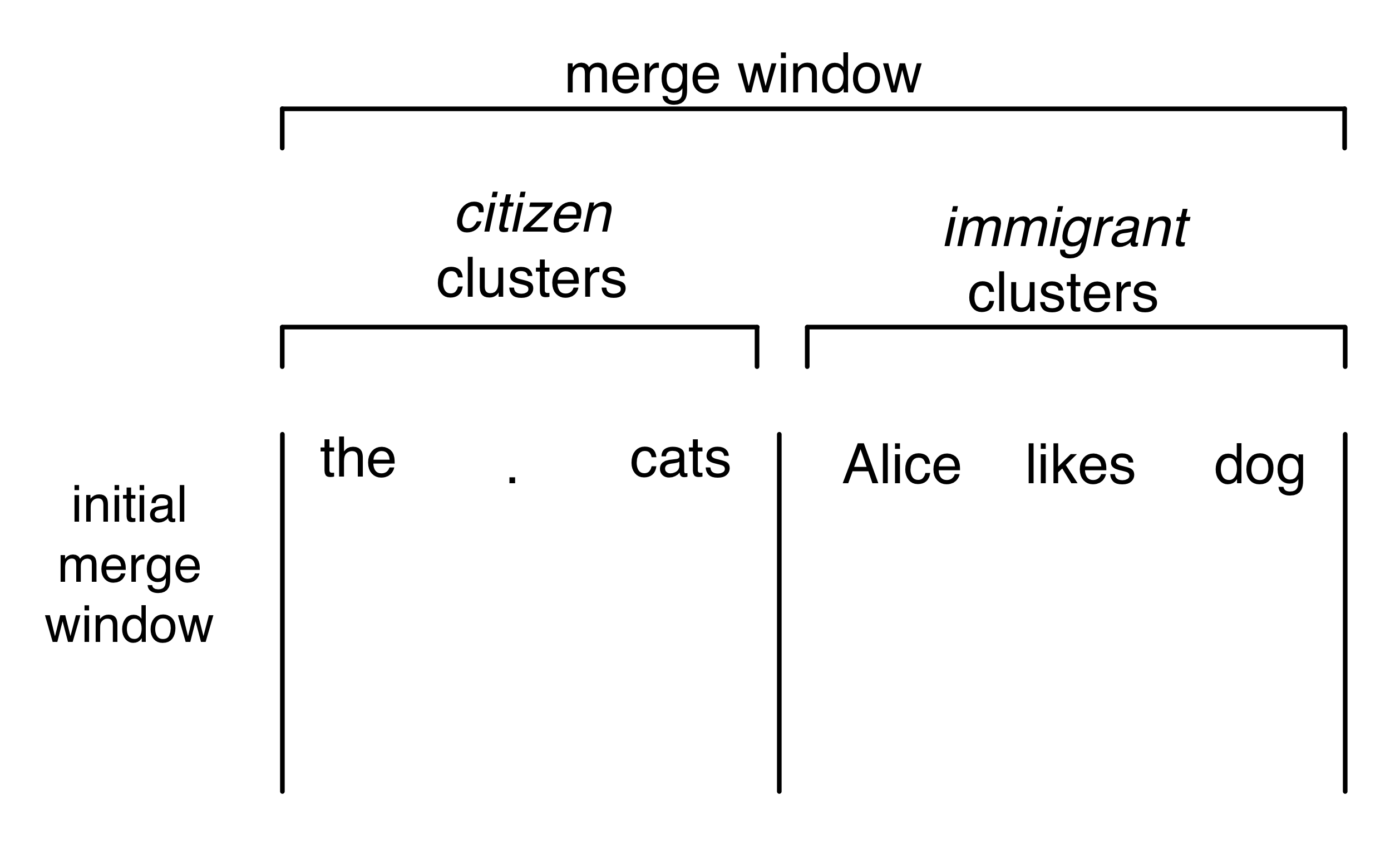}
	\caption{The conceptual split of the merge window in the \ALLSAME 
		algorithm. Any merger must contain at least one cluster from the left 
		section. The figure represents the initial merge window state when 
		starting the \ALLSAME algorithm on the corpus from Figure 
		\ref{fig:example_text}.}
	\label{fig:ALLSAME_diagram}
\end{figure}

When encountering several words with the same 
number of occurrences the algorithm should temporarily enlarge its merge window 
so as to include all these words and then temporarily change its merge 
strategy. It should no longer allow mergers between any two clusters, but 
enforce the requirement that at least one of the merged clusters was part of 
the merge window before the last mass inclusion (a \emph{citizen} cluster). 
This way, 
the greedy merge strategy is prevented from lumping together 
single-word-clusters into larger clusters that it then deals with expensively. 
The requirement also deals with the problem of choosing between words with the 
same number of occurrences, since all candidates are brought into the 
window. Figure \ref{fig:ALLSAME_diagram} illustrates the conceptual split of 
the clusters in the merge window.
After a mass inclusion, the algorithm will continue making merges, but 
no inclusions. As time passes, and the merge window 
will become smaller and there will be fewer candidates from the original 
window. When the merge window reaches size $C$, it will consist of 
two types of clusters: larger versions of those that were present in the merge 
window before the mass inclusion, and a 
group of single-word-clusters that were included as part of the mass inclusion. 
The single-word-clusters are those whose merger presented the highest 
loss in average mutual information. The single-word-clusters can now 
become ``citizens`` of the merge window, meaning that at the next mass 
inclusion they will be allowed to participate in mergers with new 
\emph{immigrants}. 
Algorithm \ref{alg:allsame} 
presents the pseudocode of the algorithm described above.

\begin{algorithm}
	\caption{The \ALLSAME algorithm}
	\label{alg:allsame}
	\algsetup{indent=2em}
	\begin{algorithmic}[1]
		\REQUIRE{S, C}
		\STATE $merge\_limit,\ current\_window\_size = initial\_inclusion (S,\ 
		C)$
		
		\COMMENT{$initial\_inclusion$ is defined in Algorithm 
		\ref{alg:allsame:initial_inclusion} on page
		\pageref{alg:allsame:initial_inclusion}}
		\FOR{$i = 0$ \TO $|V_S| - C$}
			\STATE $(C_i, C_j) \leftarrow 
			best\_merge(C,\ merge\_limit,\ current\_window\_size)$
			
			\COMMENT{$best\_merge$ is defined in Algorithm 
				\ref{alg:allsame:best_merge} on page
				\pageref{alg:allsame:best_merge}}
			\STATE $C_i \leftarrow C_{i \leftarrow j}$
			\FOR{$k = j$ \TO $current\_window\_size - 1 $}
				\STATE $C_k = C_{k+1}$
			\ENDFOR
			\IF{$current\_window\_size = C$}
				\STATE $merge\_limit \leftarrow C$
				\IF{$i + current\_window\_size \neq |V_S|$}
					\STATE $k \leftarrow i + current\_window\_size$
					\STATE $occurrences\_previous = occurrences[k]$
					\WHILE{$occurrences[k] = occurrences\_previous$}
						\STATE include word $V_S[k]$ as $C_k$
						\STATE $k = k + 1$
						\STATE $current\_window\_size = current\_window\_size + 
						1$
					\ENDWHILE
				\ENDIF
			\ELSE
				\STATE $current\_window\_size \leftarrow current\_window\_size 
				- 1$
				\IF{$i < C + 1$ \AND $j < C + 1$}
					\STATE $merge\_limit \leftarrow merge\_limit - 1$
				\ENDIF
			\ENDIF
		\ENDFOR
	\end{algorithmic}
\end{algorithm}

Line 1 of Algorithm \ref{alg:allsame} calls the \emph{initial\_inclusion} 
function described in Algorithm \ref{alg:allsame:initial_inclusion}. 
\emph{initial\_inclusion}
creates the initial clusters and merge window. This is done taking into account 
all words that have the same number of occurrences. If the words on position 
$C+1$ and $C+2$ have the same number of occurrences, both are included in the 
initial merge window as well as all other words that have the same number of 
occurrences. The $merge\_limit$ variable keeps track of the largest identifier 
of a cluster that can be allowed on the left side of a merge. Initially, this 
is set to be the identifier of the last word in the window that has the second 
largest number of occurrences (line 7 in Algorithm 
\ref{alg:allsame:initial_inclusion}).

\begin{algorithm}
	\caption{The $initial\_inclusion$ method}
	\label{alg:allsame:initial_inclusion}
	\algsetup{indent=2em}
	\begin{algorithmic}[1]
		\REQUIRE{S, C}
		\STATE $merge\_limit \leftarrow 0$
		\STATE $previous\_occurrence \leftarrow 0$
		\STATE $i \leftarrow 0$
		\LOOP
		\IF{$occurrences[i] \neq previous\_occurrence$}
		\IF{$ i \leq C + 1$}
		\STATE $merge\_limit \leftarrow i$
		\STATE $previous\_occurrence \leftarrow occurrences[i]$
		\ELSIF{$i > C+1$}
		\STATE break
		\ENDIF
		\ENDIF
		\STATE include word $i$ in window as $C_i$
		\STATE $i \leftarrow i + 1$
		\ENDLOOP
		\RETURN $merge\_limit$, $i$
	\end{algorithmic}
\end{algorithm}

Lines 2 to 19 of Algorithm \ref{alg:allsame} contain the merge loop. In line 3 
the best merge candidates are established by calling the function $best\_merge$ 
defined in Algorithm \ref{alg:allsame:best_merge}. This function iterates over 
all combinations of $i$ and $j$ that meet the requirement that $i < 
merge\_limit$ and return the best candidates. In line 4 the cluster $C_i$ is 
replaced by $C_{i \leftarrow j}$ which is the cluster resulting from the merger 
of $C_i$ and $C_j$. In lines 5 and 6 the empty hole left after the merger of 
$C_j$ is covered by moving all clusters $C_k$ with $k > j$ one position to the 
left. This strategy for covering empty holes is different from the ones used so 
far. \citet{chester2015brown} consider the collection of current clusters to be 
a set, so removals are easy. \citet{brown1992class} propose moving the last 
cluster in the clustering into position $j$. In \ALLSAME, the order of 
clusters is important as it is used when deciding which clusters can be part of 
merges.

When the merge window reaches size $C + 1$ the $merge\_limit$ variable is reset 
to the value $C$ (line 8). Lines 9 to 15 take care of including all words that 
share the next highest number of occurrences. Lines 17 to 19 update the merge 
window size and $merge\_limit$ after a merge that is not followed by an 
inclusion.

\begin{figure}[h!]
	\centering
	\includegraphics[width=0.6\textwidth,natwidth=3382,natheight=8166]{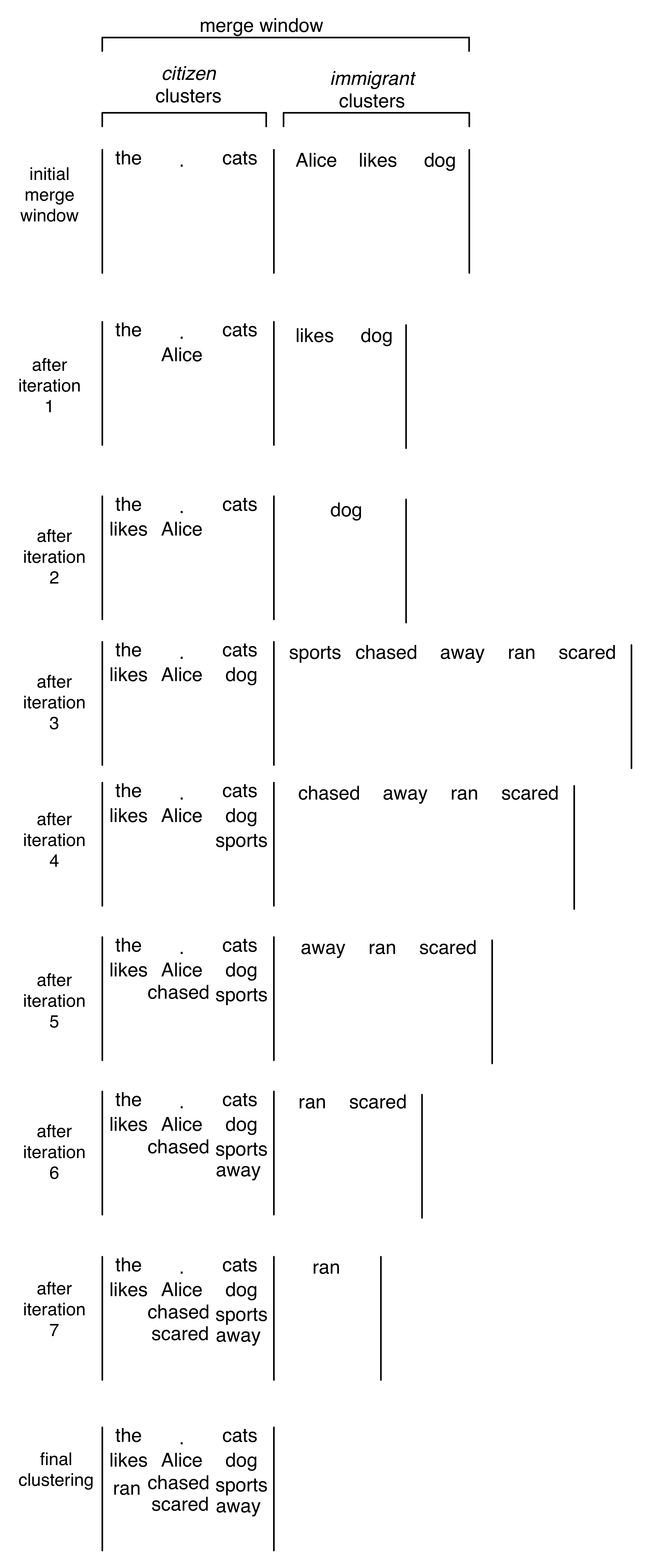}
	\caption{A step by step run of the \ALLSAME algorithm on the corpora from 
	Figure \ref{fig:example_text}. The algorithm is run with parameters $C = 3$ 
	and $w=8$.}
	\label{fig:ALLSAME_diagram_run_example}
\end{figure}

When run with $C=3$ and $w=11$, \ALLSAME will return the clustering in Figure 
\ref{fig:example_text:b} when given either the corpus in Figure 
\ref{fig:example_text:a} or the one in Figure \ref{fig:example_text:b}. A 
clustering independent of the order words appear in the corpus is what we want 
as this way, results are reproducible. In Figure 
\ref{fig:ALLSAME_diagram_run_example} I show a step by step diagram of 
\ALLSAMENOSPACE's merge decisions. One can see both at the initial merge window 
and the merge window after iteration~3, that \ALLSAME makes its merge decisions 
based on all words with the same frequency, not just on words that appear first 
in the corpus (like the windowed Brown clustering does).

\begin{algorithm}
	\caption{The $best\_merge$ method}
	\label{alg:allsame:best_merge}
	\algsetup{indent=2em}
	\begin{algorithmic}[1]
		\REQUIRE{C, $merge\_limit$, $current\_window\_size$}
		\FOR{$i = 0$ \TO $merge\_limit$}
			\FOR{$j = i + 1$ \TO $current\_window\_size$}
				\STATE calculate loss of merging $C_i$ with $C_j$
			\ENDFOR
		\ENDFOR
		\STATE find $C_i$ and $C_j$ with smallest loss
		\RETURN $C_i$, $C_j$
	\end{algorithmic}
\end{algorithm}

\section{Summary}
In this chapter I presented \RESORT and \ALLSAMENOSPACE, two modifications to 
the Brown clustering algorithm meant to address the issues described in 
Chapter~\ref{chapter_analysis}. The \RESORT algorithm re-sorts words not in the 
merge window after every $R$ merges. This allows it to make informed inclusion 
(and subsequently merge) decisions based on the amount of average mutual 
information. I have shown through examples how \RESORT is better making 
inclusion decisions than the windowed Brown algorithm. The second part of the 
chapter presented \ALLSAMENOSPACE, a modification that solves the issue of 
words with the same frequency. \ALLSAME temporarily includes all words with the 
same frequency into the merge window and then imposes some restrictions on 
which clusters can be part of mergers. By taking into account all clusters with 
the same frequency, \ALLSAME makes better merge decisions and is expected to 
outperform windowed Brown in the amount of final average mutual information.

\chapter[Empirical validation]{Empirical validation of both 
clustering and the resultant language models}
\chaptermark{Empirical validation}
\label{chapter_experiments}

In the previous chapter, I proposed two changes to the windowed Brown 
clustering algorithm: \ALLSAME (Algorithm~\ref{alg:allsame}) expands the merge window to include all 
words with the same frequency and imposes some restrictions on which 
clusters can participate in mergers. \RESORT (Algorithm~\ref{alg:resort}) uses a dynamic sorting of the 
vocabulary in order to create merge windows with the largest possible 
amount of average mutual information.

 In this chapter I present an experimental evaluation 
of the two changes. I measure cluster quality though average mutual 
information. In order to obtain some idea of real world effects of the 
various algorithm changes, I also measure performance of the class-based 
language models that result from the word clusterings.

I start by describing the data sets used in experiments (Section~\ref{sec:datasets}). I then provide an 
overview of the various metrics employed and the choice for said metrics (Section~\ref{sec:metrics}). 
Then, I describe reproducibility details of the implementation (Section~\ref{sec:implementation}).
The last section in this chapter contains all experiment results and analyses (Section~\ref{sec:results}).

\section{Data set selection and retrieval}
\label{sec:datasets}

To construct the datasets, I started with a complete dump of all English 
Wikipedia articles as of September 1st, 2015 \cite{enwiki}. The choice of 
downloading all articles from Wikipedia was motivated by 
the fact that other widely available corpora are not good representations of 
language use. Corpora based on content from social networks, such as Twitter, 
contain too many abbreviations and inconsistent spellings. On the other hand, 
widely known corpora such those consisting of content from news agencies, have 
consistent spellings, but have a restricted use of language, guided by 
corporate style guides for journalists.

Before using the Wikipedia data, I first processed the data 
dump in order to remove most of the Wikipedia-specific markup using the 
\emph{wikiextractor} script created by \citet{attardi}. 
I then wrote a Python script to remove the remaining XML and HTML tags by 
finding all words that started and ended with angled brackets, and 
removing them. Finally, I generated 4 corpora of different lengths and 
vocabularies by taking 4 subsamples at of the Wikipedia corpus. Each of 
the 4 corpora starts at a different, random, location in the 
Wikipedia corpus and has no any overlap with the others. 
Corpora $T10$, $T20$, and $T50$ were used for 
clustering. The fourth corpus, $TT$, was used to evaluate the performance of 
class based models resulting from the various clustering algorithms. The 
corpora are named using the scheme $T[number\ or\ T]$ where digits 
indicate how long the corpus is, in thousands of words. The second $T$ in 
$TT$ stands for \emph{test}. I decided to use corpora of smaller length and 
vocabulary after starting experiments on larger corpora and evaluating the 
estimated run time to completion. For example, the longest running algorithm 
would have required approximatively 300 days to complete on a corpus of length 
$1\,000\,000$ words with a vocabulary of approximatively $56\,000$ words. 
Finally, the SimpleTokenizer class from the Apache OpenNLP library was used to 
separate all words in every corpus.

I chose to generate a separate testing corpus 
in order to follow the common practice in Machine Learning of testing on 
samples that have not been used for learning. This is supposed to provide a 
better picture of how well models generalize to yet unseen data and avoid 
over-training. An 
overview of the data sets is provided in Table \ref{tab:overview_corpora}.

\begin{table}
	\centering
	\begin{tabular}{|c|c|c|}
		\hline
		\textbf{Name} &  \textbf{Text Length}      & \textbf{Vocabulary Size} \\
		\hline
		\hline
		T10  &  $10\,000$   & 2657            \\
		\hline
		T20  &  $20\,000$   & 4420            \\
		\hline
		T50  &  $50\,000$   & 8616            \\
		\hline
		TT   &  $10\,000$   & 2815            \\     
		\hline
	\end{tabular}
	\caption{Overview of the corpora used in experiments.}
	\label{tab:overview_corpora}
\end{table}

\citet{Martin1998} changed the vocabulary of their corpora by replacing 
low frequency words with a placeholder word. \citet{brown1992class} also 
reduced the vocabulary size in their work, but they do not mention the method 
used to reduce vocabulary size. In order to experiment on real-world 
corpora, I made no changes to the vocabularies of my corpora. Generally, 
vocabularies are reduced in order to get rid of words with small frequencies, 
say, 1, or 2. Besides substantially reducing the vocabulary size, this normally 
creates a more dense co-occurrence matrix. On average, this increases the 
amount of information about every word in the remaining vocabulary. There are 
also some disadvantages to reducing the vocabulary size. For one thing, 
dropping words with low frequency severely reduces the vocabulary, which leads 
to reduced language models. Secondly, the common approach for reducing 
vocabularies is to replace undesired words with a placeholder word. 
Replacing lots 
of words with a placeholder word effectively clusters all those words into a 
massive cluster of unrelated words with no information theoretic basis and no 
particular reason for semantic nor syntatic similarity. 
The new cluster will have both high 
frequency and high connectivity with other words in the vocabulary, which might 
influence mergers in both the Brown clustering algorithm, as well as in the 
Exchange Algorithm.

\section{Evaluation metrics}
\label{sec:metrics}

I evaluate my algorithms by measuring quality of clusters 
and performance of the resulting class-based language models. I measure the 
quality of clusters using Average Mutual Information (AMI), as defined in 
Equation \ref{eq:mutual_information} on page \pageref{eq:mutual_information}. 
Since higher AMI is the optimization goal of Brown clustering, the final value 
of 
AMI will show how well each algorithm has fulfilled its optimization goal. Even 
though only the final value of average mutual information matters in evaluating 
the quality of a clustering, I also measure the amount of average mutual 
information at every iteration of the clustering algorithm. This provides a 
clearer picture of how the various algorithms behave.

In order to evaluate cluster quality as part of language models, I used the 
models in conjunction with the testing corpus, $TT$. I measured 
\emph{perplexity} for comparability with related work and \emph{class 
prediction accuracy} in order to avoid some weaknesses of \emph{perplexity}.

Perplexity is a measure of both a model and the sample under test. It measures 
how well a model predicts the given sample. It is commonly used to 
measure 
performance of statistical language models (c.f., \cite{brown1992estimate, 
jelinek1977perplexity, 	jardino1994automatic, brown1992class, 
Martin1998}).  For bi-gram models like the one used in this thesis, perplexity 
is defined as~\cite{jelinek1977perplexity}:
\begin{align}
PP(W) = \sqrt[N]{\prod_{i=1}^{N}\frac{1}{\p{w_i | w_{i-1}}}}
\end{align}
where $N$ is the length of the test corpus. As its name implies, perplexity 
measures the amount of \emph{surprise} a given test sample represents for 
a 
\begin{align}
CPA = \frac{\text{\# correctly predicted classes}}{\text{\# class predictions}}
\end{align}
specific model. If the model predicts a sample perfectly (i.e. for all $w_i,\ 
w_{i-1}$, $\p{w_i | w_{i-1}} = 1$), the value under the order 
$N$ root will be 
equal 
to $N$, resulting in a perplexity of 1. However, if the model does not predict 
the sample perfectly, the perplexity value increases. So, higher perplexity 
means a model is worse at predicting a given corpus. When I measure perplexity 
in my experiments, I do not take the order $N$ root, as the computation is both 
impractical and unnecessary. Since all language models trained on the same 
corpus have the same vocabulary, relative to a the testing corpus $TT$, they 
have the same number of transitions $N$, so their perplexity values are 
directly comparable even without the order $N$ root. Note that 
differences between language models are thus exponentiated by the missing root.

Even though perplexity has been used as an evaluation metric in previous 
research (c.f. \cite{brown1992class, Martin1998}), it is not a good performance 
measure for class-based models. This is because a class-based language 
model with $C$ classes can only predict $C$ words of the total 
words in a vocabulary. Class-based language models always predict the most 
popular word in each class. Equation \ref{eq:class-based_language_models} 
defined the probability of words in a class-based language model. When used for 
prediction, all language models return only the most likely word (i.e. the one 
with the highest probability). Given the form of Equation 
\ref{eq:class-based_language_models}, this means a single word 
per class, effectively reducing the prediction vocabulary to one of size $C$ 
(number of clusters). In other words, for any word $w_{i-1}$ from class $C_{i-1}$, 
a class-based language model will always predict the same word $w_i$ from class 
$C_i$, where $w_i$ is the most often occurring word in class $C_i$. Because 
perplexity takes into account the probability of predicting 
every word in the vocabulary (including those that will not actually be 
predicted), it will always be overly positive about the performance of a class-based 
language model. There is no problem in using perplexity to evaluate 
non-class-based language models as they can predict all words in their 
vocabulary. However, I anticipate performance of all algorithms studied in this 
thesis to be weak in terms of perplexity.

In my experiments I report the perplexity of every language model, in 
order to keep in line with the currently used methods. However, I also include 
another performance metric, the \emph{Class Prediction Accuracy} (henceforth {\em CPA}). This 
measures how often the language models predicts the class containing the next 
word. Since class-based language models can be used for reducing the search 
space for regular language models by limiting the candidate words to only those 
in a class, this metric is more appropriate than perplexity.

\section{Implementation details}
\label{sec:implementation}

In order to evaluate my algorithms, I started by implementing the 
original Brown clustering algorithm. I knew from the beginning that Brown 
clustering is a slow algorithm. For that reason, I chose to implement 
everything in the \texttt{C++} programming language in order to avoid the slowdowns that 
come with using high level virtual machines such as those for the \texttt{Java} and 
\texttt{Python} programming languages.

I found out that a lot of development time is wasted if, on every run of Brown clustering, 
the computer must read the corpus, reorder words, and then construct the 
necessary data structures. In order to avoid this wasted time I created data 
structures common to all my variations and implemented separate steps for this 
operation. My implementations can read a corpus into the data structure and 
persist the data structure to a file. This is done by making use of the Boost 
Serialization library.They can take a data structure from a 
file and reorder words, and, of course, can read a data structure from a file 
and perform Brown clustering. A similar approach is used for language models, 
which use a common data structure that is saved to a file at the end of 
clustering. Testing the language models is a separate step.

In order to reduce running time I tried, to the best of my ability, to 
parallelize work so as to take advantage of multiprocessor systems. I used 
\texttt{OpenMP} to parallelize work at various locations in each implementation. For 
example, most data structure initializations are performed in parallel, using 
static scheduling of threads. Computing every hypothetical merge and 
establishing the lowest loss is performed in parallel, using dynamic scheduling 
in order to account for the unbalanced amount of work.

In order to speed up the insertions of words into the merge window I constructed 
a data structure to keep track of which words appear together in the corpus. 
This allows for quicker insertions of low frequency words. Similarly, the code 
testing language models uses cached data structures to avoid expensive searches.

Finally I used separate machines for running learning tasks and running 
evaluation tasks. All experiments were performed on a workstation with two 
Intel Xeon E5-2687W v3~\cite{xeon} and 128 GB of RAM. I ran all language model 
tests on a machine with an Intel Xeon E5-2650 and 32 GB of RAM. The choice of 
using two machines is motivated by the large run time necessary for experiments 
relative to the amount of time available for a masters' thesis. Over the course 
of this thesis, between the various experiments, I have accumulated over 500 
hours of computation time with a CPU utilization of over 80\% across all twenty cores.

\section{Results}
\label{sec:results}
In this section I report and discuss experimental results.  First (Section 
\ref{sec:evaluating_allsame}), I look at how different windowing strategies 
affect the quality of clusters and performance of resulting class-based 
language models. Second (Section \ref{sec:evaluating_brown}), I look at how the 
windowed version of Brown clustering (referred to as BROWN) performs as 
compared to the first algorithm proposed by \citeauthor{brown1992class}, the 
non-windowed Brown clustering (referred to from now on, as \BROWNNWNOSPACE), 
implemented without the dynamic programming.

I illustrate the performance of language models and the final AMI by using 
tables. The plots show AMI values as clustering progresses. These help better 
understand why and how algorithmic decisions affect the final value of average 
mutual information by showing instantaneous change in mutual information and 
the quality of each micro-decision the algorithm makes.

In the following plots, I use short names to identify the various 
combinations of algorithms, parameters and corpora. In Table 
\ref{tab:overview_corpora}, I gave an overview of the short names for corpora. 
I use the following convention for naming experiment runs: 
[\emph{algorithm\_name}].[\emph{value of C parameter}]. Where 
\emph{algorithm\_name} is one of BROWN (see \cite{chester2015brown} for a nice 
pseudocode expression of the algorithm), \ALLSAMENOSPACE (Algorithm 
\ref{alg:allsame}), \RESORT (Algorithm \ref{alg:resort}) or \emph{BROWN\_NW}. 
The first three are described in Chapters \ref{chapter_analysis} and  
\ref{chapter_improvements}. By \emph{BROWN\_NW}, I mean the version of Brown 
clustering that does not use any window but, instead, makes decisions over 
the 
entire vocabulary. This is the first algorithm proposed by 
\citet{brown1992class}.

\begin{figure}[h!]
	\centering
	\includegraphics[width=1.0\textwidth]{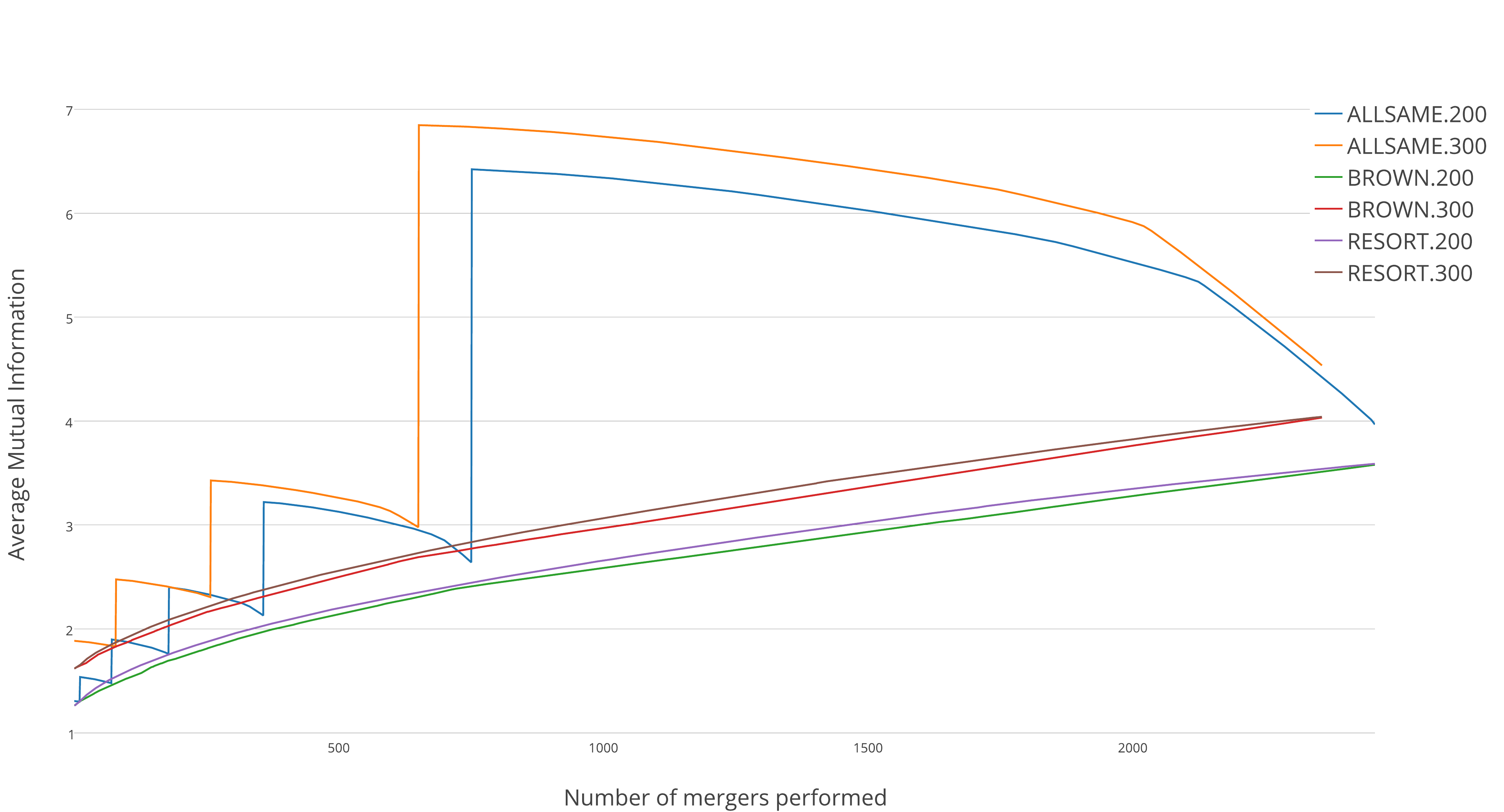}
	\caption{The average mutual information as the clustering algorithms 
		progress on the T10 corpus.}
	\label{plot:ami_progression_T10}
\end{figure}

\begin{figure}[h!]
	\centering
	\includegraphics[width=1.0\textwidth]{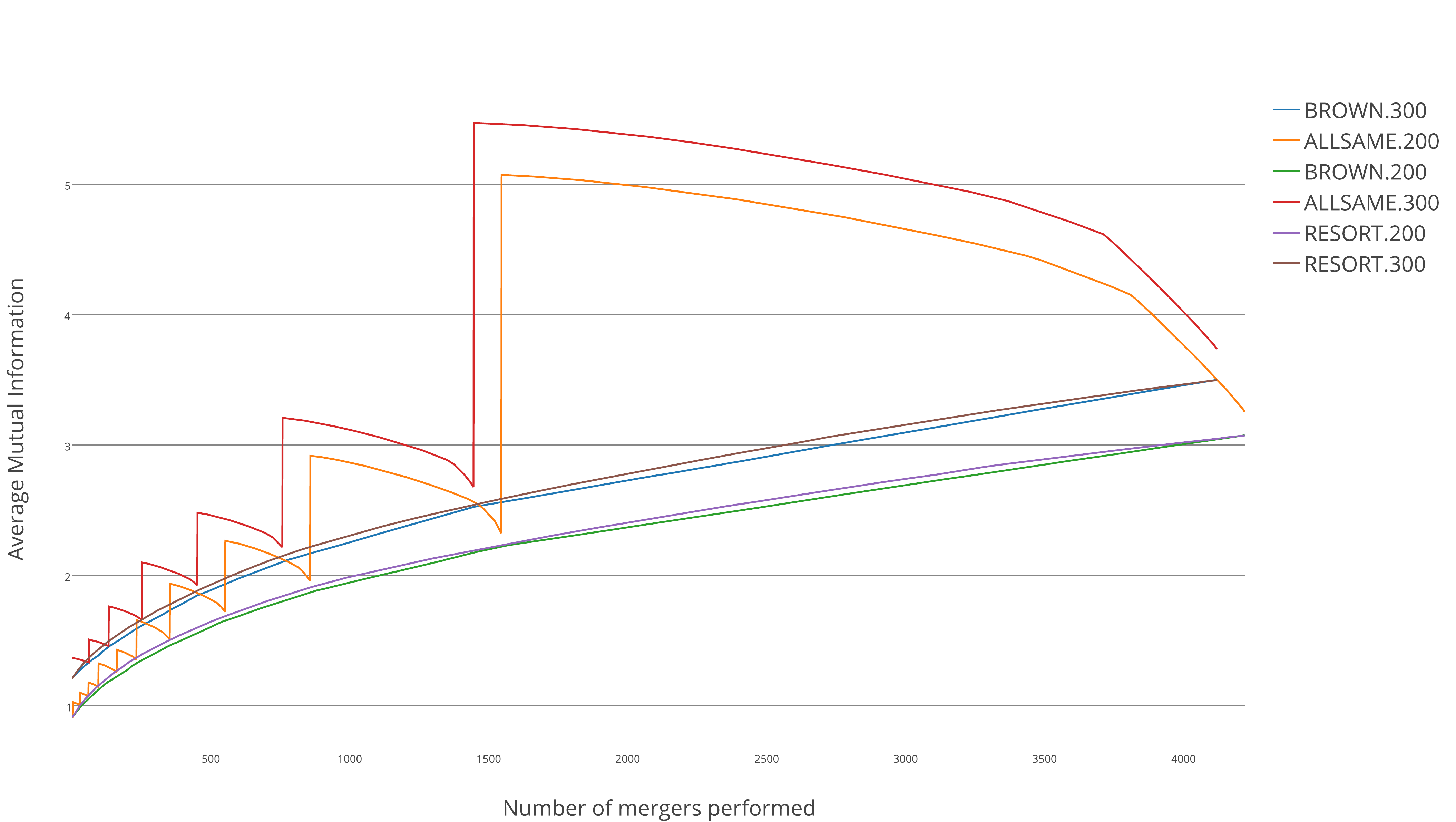}
	\caption{The average mutual information as the clustering algorithms 
		progress on the T20 corpus.}
	\label{plot:ami_progression_T20}
\end{figure}

Astute readers will notice that, in 
Figures \ref{plot:ami_progression_T10}, \ref{plot:ami_progression_T20}, 
\ref{plot:ami_progression_T50}, \ref{plot:ami_progression_NW_T10}, 
\ref{plot:ami_progression_NW_T20} 
showing 
progressions of average mutual information, runs where the $C$ parameter is set 
to $200$ have a shorter line. This is the expected behavior of all algorithms. 
Since there are $|V| - C$ merges to be performed, runs where the $C$ parameter 
is set to a higher value will perform fewer merges. The reader should focus on 
the general shape of the plots and on the $y$-coordinate of the rightmost point in each 
line, rather than where on the $x$-axis the line ends. Lines corresponding to 
runs with higher values of $C$ are expected to always be higher than those 
corresponding to lower values of $C$. This is because the Brown clustering 
algorithm is not forced to make as many expensive choices when it can assign 
words to more clusters.


\subsection{Evaluating the performance of clusters and language models 
generated with alternative windowing strategies}
\label{sec:evaluating_allsame}

This section presents the quality differences between clusters generated by the 
original Brown clustering algorithm (denoted as BROWN) and the two algorithmic 
modifications proposed in Chapter \ref{chapter_improvements}: \ALLSAMENOSPACE 
(Algorithm \ref{alg:allsame}) and \RESORT (Algorithm \ref{alg:resort}). 
\ALLSAME and \RESORT use a different windowing strategy from the windowed 
Brown algorithm. \ALLSAME includes all words with the same rank into the merge 
window and imposes some restrictions on which clusters can participate in 
mergers. \RESORT includes one word at a time into the merge window (like 
BROWN); however, it does not maintain a static ordering of words in the 
vocabulary. After every $k$ mergers, words not already included in the merge 
window are resorted by the amount of mutual information they have with words 
already in the window.

\subsubsection{Clustering quality}
\begin{figure}[h!]
	\centering
\includegraphics[width=1.0\textwidth]{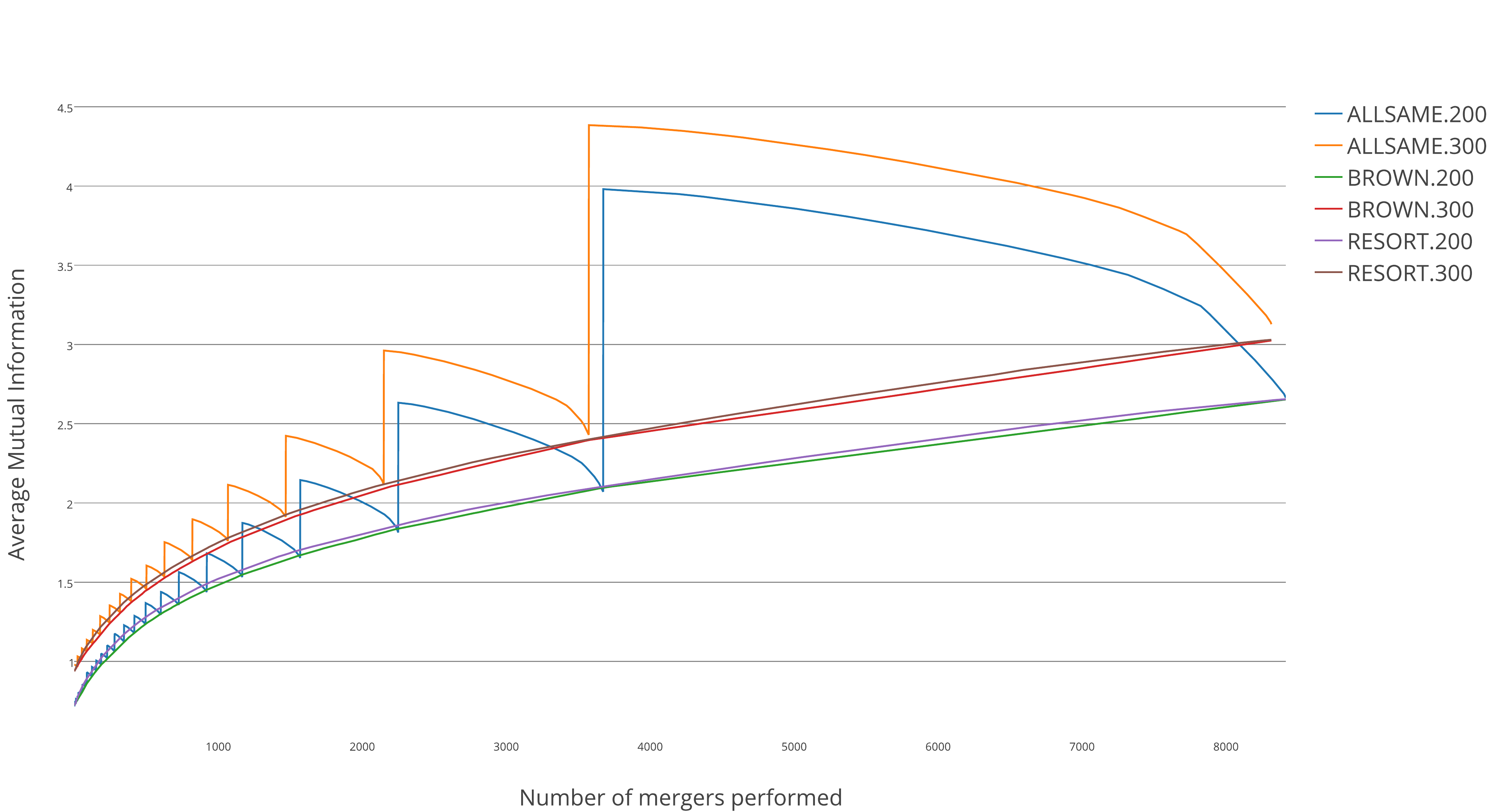}
	\caption{The average mutual information as the clustering algorithms 
		progress on the T50 corpus.}
	\label{plot:ami_progression_T50}
\end{figure}

Figures \ref{plot:ami_progression_T10}, \ref{plot:ami_progression_T20} and 
\ref{plot:ami_progression_T50} show a similar pattern over the three corpora, 
so I will discuss them together. The three plots show the amount of mutual 
information ($y$-axis) plotted against the number of mergers completed ($x$-axis). Each 
of the plots corresponds to one of the 3 learning corpora described at the 
beginning of this chapter. Each line represents the instantaneous amount of 
average mutual information. Only the final AMI is relevant in judging the 
quality of clusters. However, the progressions provide insights into the 
quality of each algorithm's decisions. In these plots, good performance is 
synonymous with an upwards trending line with a high end point. It is expected 
that the amount of average mutual information increases, as more words are 
included into the window. Every new word in the window adds $2C$ terms to the 
summation for AMI (see Equation \ref{eq:mutual_information}). A low frequency 
word increases AMI less. Because of the word's lower frequency, there are more 
clusters for which the numerator in Equation \ref{eq:mutual_information} is 
zero (no co-occurrences), leading to more zero-valued terms in the summation. 
Every merger reduces the amount of AMI as it removes $2C$ terms from the 
summation in Equation \ref{eq:mutual_information} and because it increases the 
denominators in Equation \ref{eq:mutual_information}.

The \ALLSAME algorithm has a very specific pattern to its line, which does not 
occur with the other algorithms. \ALLSAMENOSPACE's average mutual information 
line displays a \emph{saw shape} due to mass inclusions of words into the merge 
window. Every tooth in the saw corresponds to including all words with a 
specific frequency. The teeth vary in size because there are more words with a 
smaller frequency so, even though each word brings little mutual information 
into the merge window, the cumulative effect is larger. \ALLSAME gains average 
mutual information over BROWN after each saw tooth.

\ALLSAME ends by outperforming 
BROWN in 5 of the 6 runs, and it matches BROWN in the remaining (see last 
column in Table \ref{tab:performance_language_models} for the actual values). 
This is due to the fact that every time \ALLSAME encounters multiple words with 
the same frequency, it chooses its own order for merging these words based on 
their average mutual information with all words in the merge window. BROWN, on 
the other hand includes and merges same frequency words based on an 
undetermined order that is influenced by the order words appear in the original 
corpus (see the discussion in Section \ref{sec:word_order_example}). Actually 
because \ALLSAME makes better merge decisions than BROWN every time there are 
multiple words with the same frequency, \ALLSAME should be expected to perform 
better than BROWN the more words with a given frequency there are (the larger 
the saw teeth), and the more often this occurs (the more teeth there are).

The \RESORT algorithm is a little more subtle. Its average mutual information 
tends to be just slightly better than BROWN while the algorithms runs. However, 
final average mutual information of \RESORT tends to be very close to that of 
BROWN (again, see last column in Table 
\ref{tab:performance_language_models} for the actual values). This, however, 
should not be considered a null result. Experiments on much larger vocabularies 
and with different sorting strategies are necessary in order to determine 
whether \RESORT is actually not better than BROWN. And, of course, performance 
of the resulting class-based language models should also be taken into account.

\subsubsection{Performance of class-based language models}
Having looked at the AMI progression of the algorithms, I now turn to the 
performance of the class-based language models that are generated from these 
clusters. Studying Table \ref{tab:performance_language_models} in more detail, 
one notices quickly that clusterings with the highest value of average mutual 
information are not always the best performing ones (boldface). Actually, 
there is no clear winning language model either when measured by class 
prediction accuracy, or by perplexity. Nor is there any clear connection 
between vocabulary size and performance of the language models.

In 4 of the 6 setups, 
clusterings with the highest average mutual information are not the best 
performing ones, when measured by \emph{class prediction accuracy}. Similarly, 
in 5 of the 6 setups, clusters with the highest average mutual information 
do not score best in \emph{perplexity}. Furthermore, \emph{class prediction 
accuracy} often (4 out of 6 times) disagrees with \emph{perplexity} on which 
language models perform best. Another interesting aspect is that \ALLSAME 
constantly achieves higher values of average mutual information, but this 
rarely translates into the best performance when evaluating the class-based 
language models.

Actually, the overall best language model according to class prediction 
accuracy is the one generated by \RESORT on $T50$ with a $C = 300$. This fact, 
together with the observations about \ALLSAME lead to deeper questions. Why is 
it that higher AMI rarely translates into better class prediction accuracy or 
perplexity? Could it be that average mutual information is not an optimal metric to 
use as an optimisation goal in Brown clustering? The results indicate that more 
thorough investigation of AMI is necessary.

\begin{table}
	\centering
	\begin{tabular}{|c|c|c|c|c|}
		\hline
		\textbf{Corpus} & \textbf{Language model} & \textbf{CPA (\%)} & 
		\textbf{Perplexity\textsuperscript{N}} 
		& 
		\textbf{Final 
		AMI} \\
		\hline
		\specialrule{.01em}{0.7em}{0em}
		T10 & ALLSAME.200&  16.65 & $1\,392\,340$ & $\mathbf{3.88}$ 
		\\
		\hline
		T10 & BROWN.200  &  $\mathbf{17.15}$   & $684\,689$ & 3.69           
		\\
		\hline
		T10 & RESORT.200 &  16.58   & $\mathbf{664\,733}$ & 3.70  \\
		\specialrule{.1em}{0em}{0em}
		T10 & ALLSAME.300&  15.99   & $782\,654$ & $\mathbf{4.33}$\\
		\hline
		T10 & BROWN.300  &  17.06& $516\,494$ & 4.14   \\
		\hline
		T10 & RESORT.300 &  $\mathbf{17.38}$& $\mathbf{503\,967}$ & 4.15 
		\\        
		\specialrule{.01em}{0em}{.1em}
		\specialrule{.01em}{.7em}{0em}
		T20 & ALLSAME.200&  $\mathbf{17.07}$ & $6\,340\,400$ & $\mathbf{3.26}$ 
		\\
		\hline
		T20 & BROWN.200  &  15.89   & $3\,993\,800$ & 3.17           \\
		\hline
		T20 & RESORT.200 &  15.46   & $\mathbf{3\,686\,350}$ & 3.17  \\
		\specialrule{.1em}{0em}{0em}
		T20 & ALLSAME.300&  16.66   & $3\,359\,050$ & $\mathbf{3.69}$\\
		\hline
		T20 & BROWN.300  &  $\mathbf{16.85}$& $2\,057\,020$ & 3.60   \\
		\hline
		T20 & RESORT.300 &  16.62   & $\mathbf{1\,897\,490}$ & 3.59  \\        
		\specialrule{.01em}{0em}{.1em}
		\specialrule{.01em}{.7em}{0em}
		T50 & ALLSAME.200&  15.73 & $45\,033\,000$ & $\mathbf{2.73}$ \\
		\hline
		T50 & BROWN.200  &  $\mathbf{16.56}$& $\mathbf{23\,851\,800}$& 
		$\mathbf{2.73}$ 
		\\
		\hline
		T50 & RESORT.200 & 16.50& $27\,125\,700$ & $\mathbf{2.73}$  
		\\
		\specialrule{.1em}{0em}{0em}
		T50 & ALLSAME.300&  16.35   & $25\,898\,900$ & $\mathbf{3.17}$\\
		\hline
		T50 & BROWN.300  &  16.52& $\mathbf{16\,686\,700}$ & 3.10   \\
		\hline
		T50 & RESORT.300 &  $\mathbf{17.30}$& $17\,850\,400$ & 3.10  
		\\        
		\hline
	\end{tabular}
	\caption{A qualitative overview of the various algorithmic changes proposed 
	in this thesis. The best 
	results in every metric shown in boldface. Quality is measured both 
	directly on clusters through final average mutual information (the 5th 
	column) and on the resulting class-based language models. The test corpus 
	$TT$ is used to measure language model characteristics such as class 
	prediction accuracy (3rd column) and perplexity (4th column). }
	\label{tab:performance_language_models}
\end{table}

\subsection{Evaluating the idea of using a window}
\label{sec:evaluating_brown}

\citet{brown1992class} proposed Brown clustering initially without the use of a 
merge window. The non-windowed Brown clustering algorithm starts by assuming 
every word is a cluster of its own and then iteratively performs $|V|-C$ 
mergers in order to obtain $C$ clusters. There are no inclusions in this 
algorithm since all words are part of the merge window from the beginning. This 
algorithm is denoted in the following figures by \BROWNNWNOSPACE. The windowed 
version (in my experiments denoted by BROWN) was proposed by 
\citeauthor{brown1992class} in the same paper as an approximation algorithm 
that can be run faster than the non-windowed Brown clustering algorithm. The 
windowed Brown clustering algorithm starts by assuming that the first $C+1$ 
most frequent words are clusters of their own. It then iteratively performs a 
merge of two clusters and then includes the next most frequent word into the 
merge window as a cluster. When there are no more words to include into the 
merge window, one more merge is performed in order to obtain $C$ clusters.
\begin{figure}[h!]
	\centering
	\includegraphics[width=1.0\textwidth]{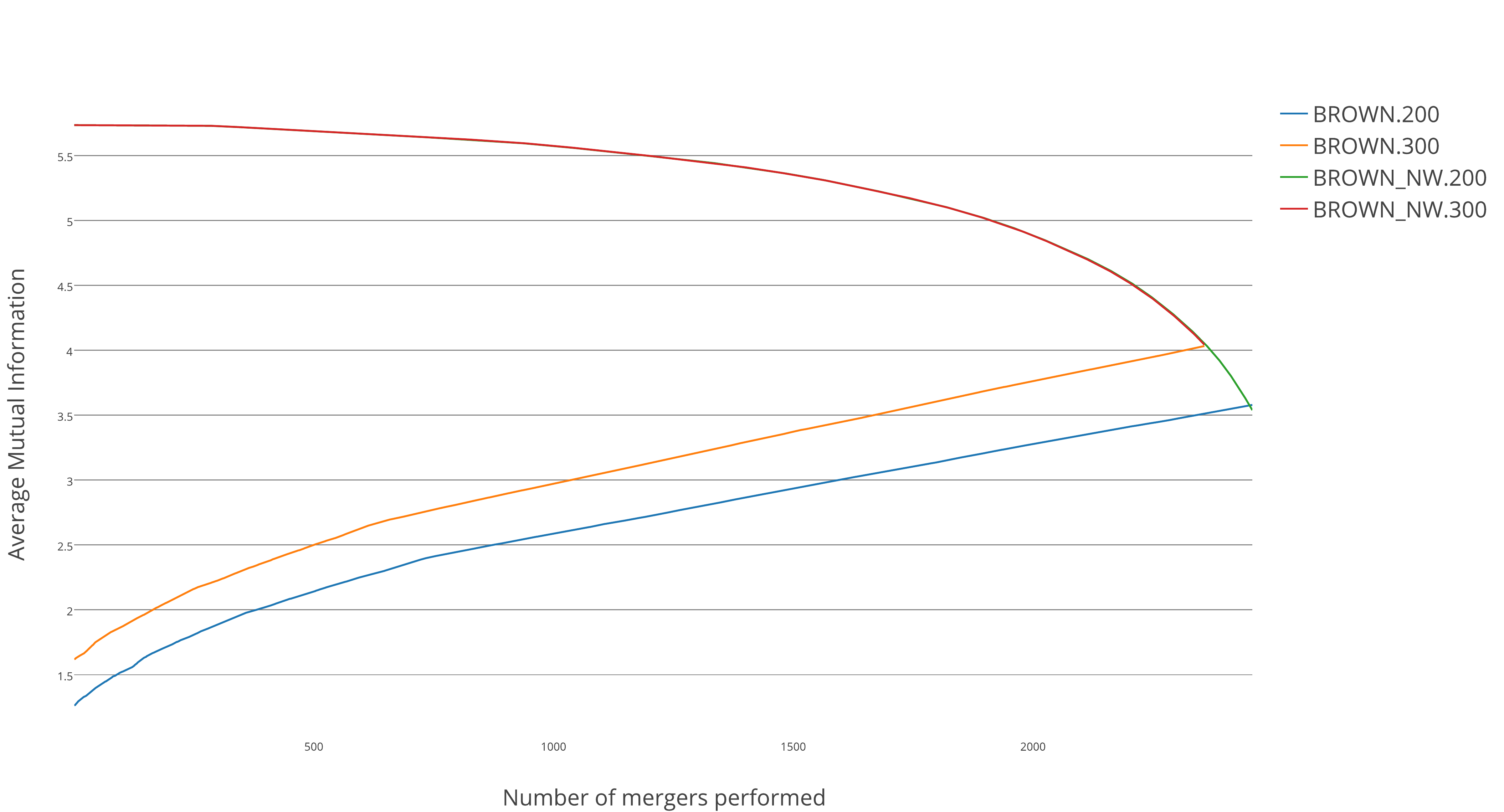}
	\caption{\BROWNNW versus BROWN on the T10 corpus.}
	\label{plot:ami_progression_NW_T10}
\end{figure}
\begin{figure}[h!]
	\centering
	\includegraphics[width=1.0\textwidth]{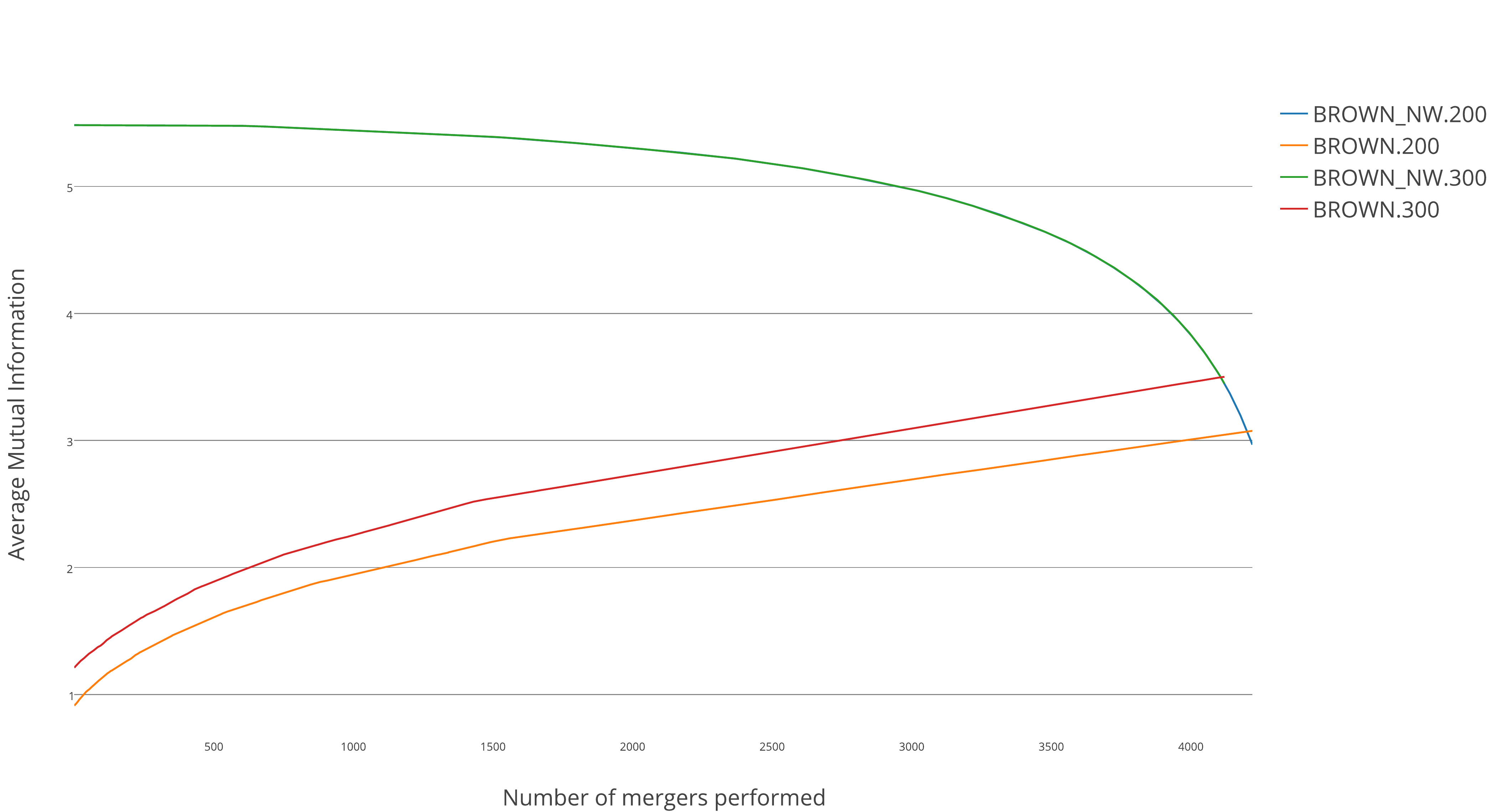}
	\caption{\BROWNNW versus BROWN on the T20 corpus.}
	\label{plot:ami_progression_NW_T20}
\end{figure}
In Section \ref{proposing_ALLSAME}, I made an argument against allowing mergers 
between any two clusters immediately after a mass inclusion. The argument was 
that when lots of low frequency, single-word clusters are introduced into the 
window, the greedy clustering strategy does not perform very well. However, if 
this is correct, it means that the non-windowed Brown clustering algorithm 
(\BROWNNWNOSPACE) should be perform worse than its approximation, 
the windowed Brown clustering algorithm. Actually, \BROWNNW should incur 
exponential increases in lost average mutual information. If we consider 
carefully the non-windowed version of the Brown clustering algorithm (which I 
refer to as \BROWNNW in all plots and tables), we realize that the algorithm 
starts by making a mass inclusion of all words in the vocabulary $V$. After 
this mass inclusion the algorithm performs $|V| - C$ mergers in order to obtain 
the $C$ clusters. In the following section I will present results of 
experiments comparing cluster quality between the original, non-windowed Brown 
clustering algorithm (referred to as \BROWNNW and the restricted, windowed, 
version (to which I will refer as BROWN). The windowed Brown clustering 
algorithm is the same algorithm as the \emph{brown-cluster} (sometimes referred 
to as \emph{wcluster}) implementation by \citet{liang2005code}, which is used 
ubiquitously in Natural Language Processing research.

Figures \ref{plot:ami_progression_NW_T10} and \ref{plot:ami_progression_NW_T20} 
follow the same design as Figures \ref{plot:ami_progression_T10}, 
\ref{plot:ami_progression_T20}, 
\ref{plot:ami_progression_T50} and 
show the progression of average mutual information of BROWN against that of 
\BROWNNWNOSPACE. As expected, \BROWNNW incurs what seems like exponential loss 
in mutual information as it makes more mergers. This is obviously not the 
desired behavior for \BROWNNWNOSPACE. Actually the difference in average mutual 
information between BROWN and \BROWNNW seems to increase as the corpus size 
increases. In order to confirm this, experiments on larger corpora should be 
executed. Unfortunately, I cannot show the results of BROWN against \BROWNNW on 
the $T50$ corpus as the large computational complexity of these experiments 
prevented them from being finished before the deadline for this thesis. 

Table \ref{tab:lang_model_NW} shows the language model performance metric and 
average mutual information of \BROWNNWNOSPACE. For easier comparison I have 
included the performance numbers for BROWN from Table 
\ref{tab:performance_language_models}. As suggested earlier in this 
subsection, BROWN achieves a higher average mutual information than \BROWNNWNOSPACE. This 
happens on both corpora. However, the situation is quite mixed when it comes 
to performance of language models. In 3 of the 4 setups, BROWN achieves better 
class prediction accuracy than \BROWNNWNOSPACE. The situation is different when 
considering the values for perplexity. Similar to Table 
\ref{tab:performance_language_models}, good average mutual information or class 
prediction accuracy are not necessarily indicators of good perplexity.

\begin{table}
	\centering
	\begin{tabular}{|c|c|c|c|c|}
		\hline
		\textbf{Corpus} & \textbf{Language model} & \textbf{CPA (\%)} & 
		\textbf{Perplexity\textsuperscript{N}} 
		& 
		\textbf{Final 
			AMI} \\
		\hline
		\specialrule{.01em}{0.7em}{0em}
		T10 & BROWN.200  &  17.15   & $684\,689$ & 3.69 \\
		\hline
		T10 & \BROWNNWNOSPACE.200 & 17.06 & $901\,309$ & 3.62 \\
		\hline
		T10 & BROWN.300  &  17.06& $516\,494$ & 4.14   \\
		\hline
		T10 & \BROWNNWNOSPACE.300 & 16.81 & $486\,449$ & 4.11 \\

		\specialrule{.01em}{0em}{.1em}
		\specialrule{.01em}{.7em}{0em}
		T20 & BROWN.200  &  15.89   & $3\,993\,800$ & 3.17           \\
		\hline
		T20 & \BROWNNWNOSPACE.200 & 17.36 & $3\,890\,810$ & 3.03 \\
		\hline
		T20 & BROWN.300  &  16.85& $2\,057\,020$ & 3.60   \\
		\hline
		T20 & \BROWNNWNOSPACE.300 & 16.64 & $1\,745\,120$ & 3.51 \\
		\hline
	\end{tabular}
	\caption{A qualitative comparison between using a window (BROWN) and not 
	having a window, equivalent to running clustering over the entire corpus 
	(\BROWNNWNOSPACE). Quality is measured both directly on clusters through 
	final average mutual information (the 5th column) and on the resulting 
	class-based language models. The test corpus $TT$ is used to measure 
	language model characteristics such as class prediction accuracy (3rd 
	column) and perplexity (4th column). Results for the $T50$ corpus are 
	omitted as the experiments on such a large corpus take too long to 
	complete.}
	\label{tab:lang_model_NW}
\end{table}

\subsection{Summary}

In this chapter I showed and discussed results of running experiments to 
measure and compare the performance of my proposed algorithmic changes \ALLSAME 
and \RESORTNOSPACE. I also discussed results of experiments trying to confirm the 
counter-intuitive idea that the non-windowed Brown clustering algorithm 
(\BROWNNWNOSPACE) performs worse than the windowed version of Brown clustering 
(BROWN).

While I could show that \ALLSAME can achieve better average mutual information 
than windowed Brown (BROWN), that result did not translate into better 
performance of the corresponding class-based language models. And while \RESORT 
did not manage to achieve better values of average mutual information than 
BROWN, it did achieve better language model performance in some of the 
experiments. This behavior of \RESORT suggests that the idea of dynamic word 
ordering should not be dismissed, but that different sorting criteria need to 
be investigated.

I managed to show that \BROWNNW achieves lower average mutual information than 
windowed Brown; however, that did not always translate into lower performance 
of the associated class-based language models. The lack of strong correlations, 
in both sets of experiments, between the performance of class-based language 
models and average mutual information raises some questions about the 
effectiveness of average mutual information as an optimization goal for Brown 
clustering.

\chapter{Conclusion and Future Work}
\label{chapter_conclusion}
\section{Conclusion}
Brown clustering is a hierarchical hard clustering algorithm designed to 
allocate words to clusters based on their usage in a learning corpus. Brown 
clustering has applications in the creation of class-based language models 
and 
for providing features for Natural Language Processing tasks.

Even though {\em twenty-three years} have passed since Brown clustering 
was proposed, not much 
research has been done towards improving the quality of resulting 
clusters, in spite of the fact that one of the two publicly available 
implementations has seen wide use in the research community.

In this thesis I have performed a close analysis of the Brown clustering 
algorithm to reveal some weaknesses. I have also provided two heuristics that 
address some of the shortcomings present in the original algorithm. I 
introduced \ALLSAMENOSPACE, a modification of the windowed Brown clustering 
algorithm to solve the question of which of the words with same frequency 
should be included first into the merge window. \ALLSAME\ includes all words 
with the same frequency into the merge window, but enforces some restrictions 
as to which clusters are allowed to participate in mergers. My second proposal, 
\RESORTNOSPACE, challenges the idea of a static word order. \RESORT reorders words not 
already included in the window after every $k$ merges using the amount of 
mutual information with words already in the window as a differentiating 
dimension.

My experiments have shown that \ALLSAME does consistently achieve better values 
of average mutual information. However, they have also shown that higher values 
of AMI are not strong indicators for high performing language models. Actually, 
the overall best performing language model was created by \RESORTNOSPACE, which did 
not generally achieve the best AMI.

\section{Future work}

One of the major difficulties in running experiments for this thesis has 
been the long running times of Brown clustering which have prevented me from 
experimenting over larger data sets. My implementation makes use of CPU 
parallelism in order to speed up computations. Radically different 
implementations of the algorithm should be researched in order to make better 
use of parallelism capabilities already present in wide-spread CPUs and GPUs 
(graphics processing units). Another direction to follow is finding ways to 
distribute the algorithm so that is can be run on multiple computers at the 
same time. Similar research has been done on distributing the exchange 
algorithm \cite{distributedExchange}, but not on Brown clustering.

I would also like to do some follow-up research on understanding the interplay 
between average mutual information and clustering quality. As my experiments 
have shown, higher values of average mutual information do not always 
lead to 
better language models. However, my experiments have not explored whether this 
is also the case for other uses of Brown clusters, such as the Natural 
Language 
Processing Tasks mentioned in Section \ref{sec:applications_brown}. Actually, 
to my knowledge, there is no research on whether Brown clusters that perform 
well as classes for class-based language models are also good clusters 
for Natural Language Processing. Evaluation of cluster quality is a topic of 
interest in clustering research as clustering has generally proven difficult to 
evaluate.

Another research avenue I would like to pursue is incorporating external 
knowledge into the clustering algorithm. In his survey, \cite[]{twodecades} 
makes an argument for linguistic information being incorporated into 
statistical language models. Indeed, \citet{Martin1998} tried to use parts of 
speech in order to initialize the Exchange Algorithm (see Section 
\ref{sec:rel:initialization}). I believe it might be interesting to explore the 
idea of assisting Brown clustering not just with linguistic information, but 
also with domain knowledge, or maybe even information from other 
clusterings.

\backmatter
\printbibliography
\addcontentsline{toc}{chapter}{Bibliography}
\end{document}